\documentclass{article}

\usepackage{microtype}
\usepackage{graphicx}
\usepackage{subfigure}
\usepackage{booktabs} 

\usepackage{hyperref}

\usepackage[arxiv]{icml2025}

\usepackage{amsmath}
\usepackage{amssymb}
\usepackage{mathtools}
\usepackage{amsthm}

\usepackage{makecell}

\usepackage[capitalize,noabbrev]{cleveref}

\usepackage{soul}

\usepackage{multirow}
\usepackage{tikz}
\def\checkmark{\tikz\fill[scale=0.4](0,.35) -- (.25,0) -- (1,.7) -- (.25,.15) -- cycle;} 

\usepackage{pifont}
\usepackage{enumitem}
\setlist[itemize]{leftmargin=0pt, labelsep=5pt, itemsep=0em, topsep=0em, parsep=0em}
\usepackage{xspace}
\newcommand{\MET}{{DocVXQA}\xspace}

\theoremstyle{plain}

\theoremstyle{definition}

\theoremstyle{remark}

\usepackage[textsize=tiny]{todonotes}

\icmltitlerunning{\MET: Context-Aware Visual Explanations for Document Question Answering}

\begin{document}

\twocolumn[
\icmltitle{\MET: Context-Aware Visual Explanations for Document Question Answering}



\icmlsetsymbol{equal}{*}

\begin{icmlauthorlist}
\icmlauthor{Mohamed Ali Souibgui}{equal,cvc}
\icmlauthor{Changkyu Choi}{equal,UiT}
\icmlauthor{Andrey Barsky}{cvc}
\icmlauthor{Kangsoo Jung}{Inria}
\icmlauthor{Ernest Valveny}{cvc}
\icmlauthor{Dimosthenis Karatzas}{cvc}
\end{icmlauthorlist}

\icmlaffiliation{cvc}{Computer Vision Center, Universitat Autònoma de Barcelona, Spain}
\icmlaffiliation{UiT}{UiT The Arctic University of Norway, Norway}
\icmlaffiliation{Inria}{Inria, France}

\icmlcorrespondingauthor{Mohamed Ali Souibgui}{msouibgui@cvc.uab.cat}

\icmlkeywords{Document Intelligence, Explainability, Self-explainable Deep Learning, Vision and Language}

\vskip 0.3in
]



\printAffiliationsAndNotice{\icmlEqualContribution} 

\begin{abstract}
We propose \textbf{\MET}, a novel framework for visually self-explainable document question answering. The framework is designed not only to produce accurate answers to questions but also to learn visual heatmaps that highlight contextually critical regions, thereby offering interpretable justifications for the model's decisions. 
To integrate explanations into the learning process, we quantitatively formulate explainability principles as explicit learning objectives.
Unlike conventional methods that emphasize only the regions pertinent to the answer, our framework delivers explanations that are \textit{contextually sufficient} while remaining \textit{representation-efficient}. This fosters user trust while achieving a balance between predictive performance and interpretability in DocVQA applications. Extensive experiments, including human evaluation, provide strong evidence supporting the effectiveness of our method. 
The code is available at \url{https://github.com/dali92002/DocVXQA}.
\end{abstract}

\section{Introduction}
\label{sec:itroduction}
\begin{figure}[t]
\vskip 0.2in
\begin{center}
\begin{tabular}{c}
     \includegraphics[width=0.95\linewidth]{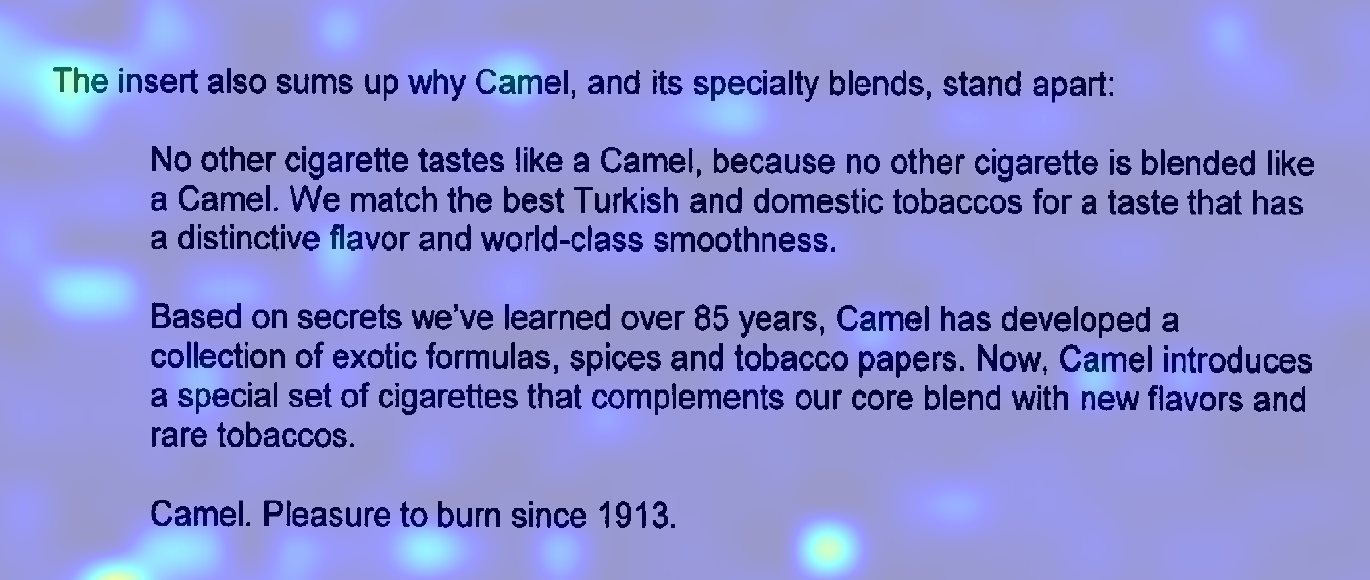}  \\
     \includegraphics[width=0.95\linewidth]{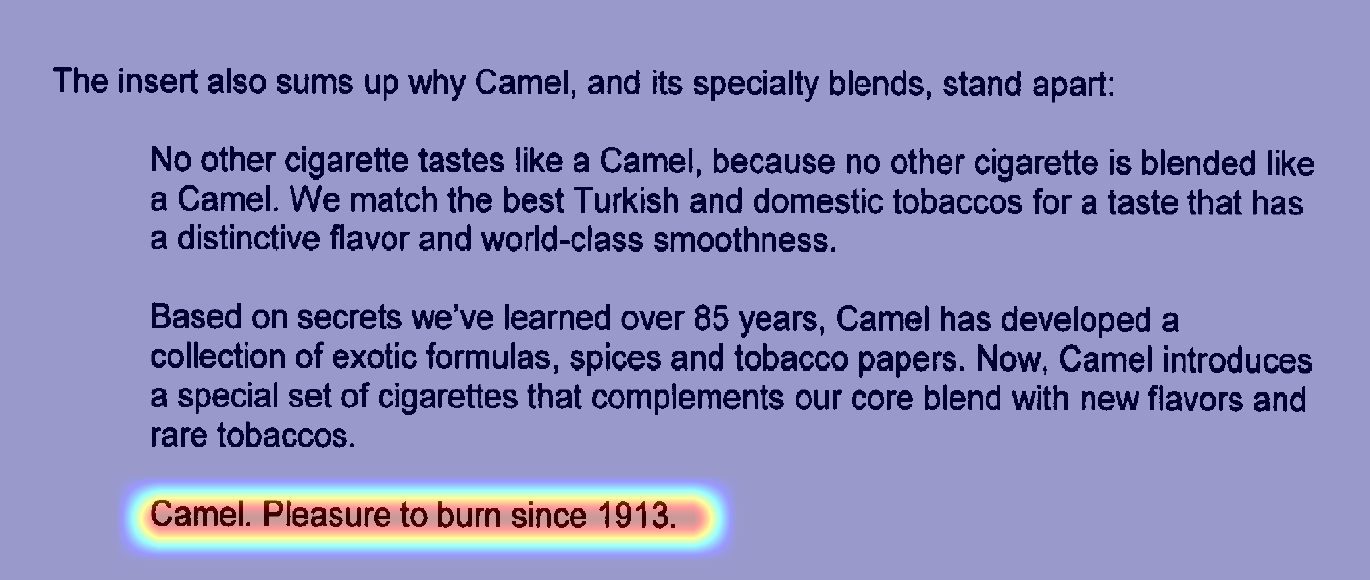}\\
     
\end{tabular}

\caption{An illustration of the relevant regions in a DocVQA model (highlighted zones), produced by extracting the raw attention maps from the last layer (top) and by using our method (bottom) for the question ``~`Pleasure to burn since 1913', Which cigarette's tagline is this?''. Here, the answer given correctly by the model is ``Camel''.}
\label{fig:teaser}
\end{center}
\vskip -0.2in
\end{figure}

Document visual question answering (DocVQA) is essential for automatically extracting information from visually complex, multi-modal documents including invoices, reports, and contracts~\cite{appalaraju2024docformerv2, blau2024gram, tito2023hierarchical, mathew2021docvqa}. This task requires interpreting the question and generating an answer through natural language, given a specific document. Recently, DocVQA models have increasingly leveraged large vision-language models for their ability to process both textual and visual modalities at scale \cite{wang2024qwen2, zhao2024evaluating, rasheed2024glamm, parashar2024neglected}. 

Despite these promising advances in terms of utility, DocVQA relies on large, opaque neural network architectures that function as ``black boxes,'' offering limited transparency into how they generate answers. This lack of transparency is particularly concerning in high-stakes domains such as finance \cite{wu2023bloomberggpt}, healthcare \cite{li2024llava}, and law \cite{abdallah2023exploring}, where trust in AI-driven decision-making is crucial.
In addition, existing explainable methods for DocVQA primarily rely on visualizing attention maps to highlight relevant regions influencing the model’s predictions. However,  as illustrated in \cref{fig:teaser} (above), these attention-based explanations are often noisy or imprecise, failing to justify the reasoning behind the answer. Instead of offering clear insights, they tend to broadly highlight all occurrences of the answer without contextualizing why it was selected.

To address this limitation and enhance the transparency of DocVQA models, we propose a novel \textit{self-explainable} framework called \textbf{\MET}, which learns to provide context-aware explanations to the answers in the form of relevance maps.
Note that self-explainable models generate interpretable outputs by design, in contrast to post-hoc approaches that provide explanations retrospectively~\cite{mollerfinding, choi2024dib, gautam2023looks, gautam2022protovae}.
While these methods have been explored from a methodological perspective, their application in the context of DocVQA remains largely unexplored. To the best of our knowledge, our framework is the first self-explainable DocVQA model designed to generate visual explanations that align with textual answers. This approach enhances both transparency and user trust in model predictions.

To enhance transparency without disrupting established DocVQA models, we propose a strategy that learns generating explanations while maintaining compatibility with available pretrained models.
We build upon the widely used Pix2Struct \cite{lee2023pix2struct} model, trained for DocVQA task. With minimal alterations to the existing architecture--thus avoiding significant impacts on performance or the need for extensive retraining--our method first learns a mask over the document image as an explanation, and then uses the masked image as an input to the same network to answer the question. This sequential learning process ensures that the generated explanations directly contribute to the answer.

The core contribution of our \MET lies in learning an effective mask designed to enhance human understanding. Grounded in a philosophical foundation of good explanation~\cite{choi2024dib, sokol2020explainability}, our approach aims to develop a method that learns mask representations that are both \textit{contextually sufficient} and \textit{representation-efficient}. To achieve this, we frame the trade-off between these objectives within the information bottleneck principle~\cite{tishby2015deep}, ensuring a balance between retaining relevant contextual information and minimizing redundancy in the representation.
Additionally, to prevent overfitting in mask learning and to enhance the generalizability of the explanations, we integrate additional pretrained models~\cite{faysse2024colpali, yu2024visrag}, which infer the mask representation as a prior and interact with the main model to refine the explanation learning process, ensuring robust and reliable outcomes. This approach offers more precise and contextually relevant justifications. Notably, although our focus is on Pix2Struct~\cite{lee2023pix2struct}, our method is inherently model-agnostic, allowing seamless integration with various DocVQA architectures.


Our main contributions are as follows:
(1) We introduce \MET, the first self-explainable DocVQA framework that learns to generate visually grounded context-aware explanations along with answer predictions.
(2) We formalize explainability as an explicit learning objective with the information bottleneck principle.
(3) We make our design model-agnostic and lightweight, thus, compatible with existing pretrained DocVQA models and requiring minimal architectural changes.
(4) We conduct extensive experiments, including human evaluations, to validate the effectiveness of the proposed approach.
\section{Related Work}
\label{sec:relatedwork}
\subsection{Document visual question answering} 

Typical DocVQA models are reaching good performances using  visual and textual features within transformer-based encoder-decoder architectures to generate answers~\cite{appalaraju2024docformerv2, tito2023hierarchical, powalski2021going, xu2020layoutlm}. In this setup, text is extrtacted with OCR systems.
However, despite their strong best-case performance, these methods are prone to errors introduced by OCR inaccuracies and face difficulty in domains that are more reliant on visual features.

In contrast, OCR-free end-to-end models \cite{lee2023pix2struct, aggarwal2023dublin, kim2022ocr, davis2022end, kim2021donut}  predict answers directly from document images, utilizing pre-training objectives to interpret text without relying on OCR. More recently, Large Vision-Language Models (VLMs) like GPT-4 \cite{achiam2023gpt} set new benchmarks in DocVQA performance, but their high computational requirements and closed-source nature limit their accessibility. There has also been a gradual emergence of smaller, open source models trained with instruction tuning \cite{wang2024qwen2, zhang2023llavar,ye2023ureader}, but all these models fall short in providing explanations that support the reason behind their predictions.
Among the OCR-free models, Pix2Struct~\cite{lee2023pix2struct} is designed for visually-situated language understanding tasks. It uses a pretraining strategy called screenshot parsing, converting masked web page screenshots into simplified HTML representations. 
The model incorporates variable-resolution input, preserving the aspect ratios for diverse document layouts.
Thus, avoiding the significant reduction in original image resolution, which make it successful in performing DocVQA without requiring external OCR systems. 




\subsection{Explainability in vision and language}
Current vision-and-language model explanations rely heavily on self-attention mechanisms to generate relevance maps, highlighting the regions that contribute to the model's decision \cite{bousselham2024grounding, chefer2021transformer, chefer2021generic}. However, attention maps are often noisy \cite{abnar2020quantifying}, making it challenging to extract meaningful insights into the model's reasoning.  Recently, in the domain of DocVQA, new approaches have been introduced to provide more fine-grained grounding by explicitly localizing answer regions \cite{mohammadshirazi2024dlava, zhou2024doge}. These methods, however, require annotated answer locations during training, leading to the development of new datasets containing such annotations \cite{giovannini2025boundingdocs}. While effective in the case of extractive DocVQA tasks (where the answer string is found explicitly in the document), these approaches introduce a significant annotation cost and remain limited in their ability to capture the context that explain why an answer is correct. 
Consequently, there remains a need for explanation techniques that not only identify specific answer locations but also provide a reasoning context beyond simple spatial answer grounding.

\begin{figure*}[t!]
\vskip 0.2in
\begin{center}
\centerline{\includegraphics[width=0.78\linewidth]{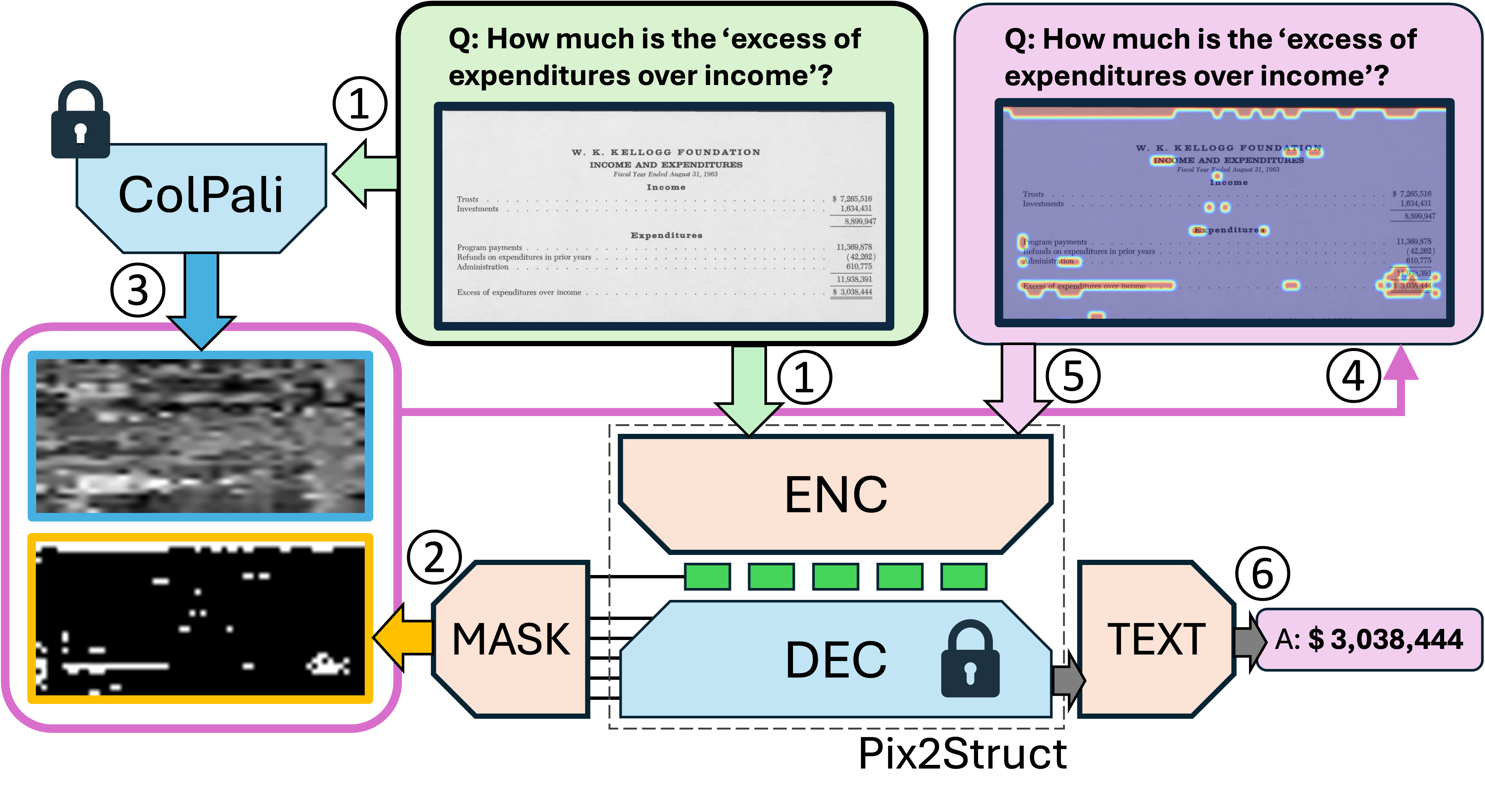}}
\caption{\textbf{Overview of the proposed \MET framework:}
\textbf{\ding{172}} The input question and the full document image are sent to both the pretrained Pix2Struct ENC-DEC and a pretrained ColPali model. \textbf{\ding{173}} The mask head (MASK) generates a learnable mask based on the decoder output and positional embeddings. \textbf{\ding{174}} ColPali provides a mask prior, highlighting the relevant regions of the input document in relation to the question. \textbf{\ding{175}} The learnable mask, guided by the mask prior, is combined with the original document to create the masked image, where only the highlighted parts are kept visible. \textbf{\ding{176}} The masked image is processed through the Pix2Struct network. \textbf{\ding{177}} The text head (TEXT) predicts the answer to the question based on the masked input.}
\label{fig:architecture}
\end{center}
\vskip -0.2in
\end{figure*}
\subsection{Learning to explain}
Model explainability has gained significant attention in recent years, aiming to provide insights into neural network decision-making \cite{covert2021explaining}. A common approach is to generate saliency maps that highlight input regions relevant to predictions \cite{schulz2020restricting, selvaraju2017grad}. In this context, the saliency map is a set of relevance scores over input pixels, intended to reveal the model’s reasoning. Earlier explainability methods provide post-hoc explanations, i.e., get explanation without changing the trained model weights or architecture~\cite{shrikumar2017learning, selvaraju2017grad, ribeiro2016should, bach2015pixel}. Gradient-based methods such as Grad-CAM \cite{selvaraju2017grad}, LRP \cite{bach2015pixel}, and DeepLIFT \cite{shrikumar2017learning} calculate saliency maps through backpropagation of model output gradients to the input or feature space. While these methods are easy to implement, they often suffer from gradient shattering, leading to noisy and imprecise explanations.
Another line of post-hoc explanation research is perturbation-based methods. This approach observes output changes by processing a set of perturbed images, each occluding different regions of the original image. Examples include LIME \cite{ribeiro2016should}, Occlusion \cite{zeiler2014visualizing}, and RISE \cite{petsiuk2018rise}. 
While these methods often produce more reliable attributions, their effectiveness is constrained by the high computational cost of sampling a sufficient number of meaningful perturbations.

Unlike post-hoc approaches, self-explainable methods \cite{choi2024dib, mollerfinding, rudin2019stop, alvarez2018towards} embed explainability directly into the training process, ensuring that models learn to generate interpretable explanations alongside predictions. Instead of relying on external attribution techniques, these models are designed to provide built-in reasoning. Among the various approaches to self-explainable deep learning methods, two main strategies stand out: the integration of mathematically formulated explainability principles into the objective function \cite{bang2021explaining, rudin2019stop, alvarez2018towards}, and architectural adjustments of the network to enhance transparency \cite{gautam2023looks, gautam2022protovae, alvarez2018towards}. 
These approaches mark a shift from the traditionally opaque 'black-box' neural networks to a more transparent and understandable framework.

\subsection{Information theoretic learning}
Information theory provides a robust foundation for quantifying data representation through metrics such as entropy, divergence, and mutual information~\cite{principe2010information, shannon1948mathematical}. 
Building on this foundation, recent learning systems have embraced these principles~\cite{skean2025frossl, yucauchy2024, yu2019multivariate, tishby2015deep} and evolved through various innovative approaches~\cite{jenssenmap2024, choi2024dib}.

\section{Method}
\label{sec:method}
\subsection{Visually self-explainable DocVQA: Enhancing context awareness through a learnable mask}



We propose \MET, a novel framework for visually self-explainable document question answering. At a high level, this framework provides not only the answer to a question but also a visual explanation in the form of a relevance map~\cite{muller2024explainable, zhang2024opti}. 
The core innovation of our method lies in learning the relevance map-based explanations while simultaneously improving their reliability. 
To achieve this, we quantitatively formalize philosophically defined explainability principles~\cite{sokol2020explainability} and integrate them into a unified objective function, enabling end-to-end training.
Our framework builds upon the widely recognized Pix2Struct architecture~\cite{lee2023pix2struct}, adapting it effectively for this domain.

\paragraph{Minimality-sufficiency trade-off} In formulating explainability principles, \MET draws inspiration from the information bottleneck (IB) framework~\cite{tishby2015deep}, which seeks to construct a compact representation $T$ of input data $X$ that preserves only the information essential for predicting a target variable $Y$. This framework balances the trade-off between reducing the mutual information $I(X; T)$ to compress irrelevant data and maximizing $I(T; Y)$ to retain predictive power.
To implement self-explainability within the conventional DocVQA setting, we strategically define the bottleneck representation $T$ as the masked input, $T = X \odot M$, where $\odot$ represents the Hadamard product that masks out irrelevant information within $X$, and $M$ is a learnable mask.
 This formulation provides a self-explaining mask in the form of region discovery, highlighting the most relevant regions of the input~\cite{choi2024dib, zhmoginov2021information}.
With these terms, the objective function $\mathcal{L_{IB}}$ is expressed as:
\begin{equation}
\label{eq:IB_basic}
\min \mathcal{L}_{IB} = \beta I(X; X \odot M) - I(X \odot M; Y),
\end{equation}
with hyperparameter $\beta$ balancing the contrasting terms.

These two terms reflect the philosophical foundation of explanation, and the principles that define what an effective explanation should be~\cite{choi2024dib, sokol2020explainability}. We frame these principles as \textit{minimality} and \textit{sufficiency}, using them as learning criteria to guide the model as direct learning signals. \textit{Minimality} emphasizes presenting only the most pertinent information for understanding a model's decision, while \textit{sufficiency} ensures that explanations remain consistent and contribute decisively to predicting a correct answer.

Figure~\ref{fig:architecture} provides an overview of our framework. 
We build on a pretrained Pix2Struct model as the baseline, updating the encoder, the newly added mask head, and the text head to enable the learning of a self-explainable mask representation. 
When a question and a document image are provided as inputs, the model first predicts a mask, introducing the minimality term $I(X; X \odot M)$ to encourage the removal of irrelevant information. 
The masked input is then processed by the model, together with the same question, to predict the answer based on the masked representation, captured by $I(X \odot M; Y)$. 
Note that the decoder of the Pix2Struct model remains frozen to ensure consistent text prediction. 
As shown in the figure, our framework also incorporates a pretrained ColPali model~\cite{faysse2024colpali}. The details of its role and integration are discussed in Sec.~\ref{sec:colpali}.

\subsection{Learning sufficient explanation via cross-entropy}
\label{sec:ce}
Upon learning the bottleneck representation, the mask $M$ must forward contextually sufficient information to the model to answer the question by maximizing the mutual information $I(X \odot M; Y)$. Following~\cite{amjad2019learning}, we approximate this by minimizing the cross-entropy loss $CE(\hat{Y}, Y)$,
\begin{equation}
\begin{aligned}
\label{eq:mi}
    \max I(X \odot M; Y) = \underbrace{H(Y)}_{\text{constant}} - H(Y \mid X \odot M) &\\
    \iff \min H(Y \mid X \odot M) \simeq CE(Y; \hat{Y})&,\\
\end{aligned}
\end{equation}
where $H(Y)$ and $H(Y \mid X \odot M)$ are the marginal and conditional entropies of $Y$, respectively. $\hat{Y}$ denotes the model prediction. Note that $H(Y)$ is constant because the label $Y$ is fixed.




\subsection{Learning minimal explanation via continuity loss with L1 norm}
\label{sec:l1}
Minimality is achieved by reducing the mutual information between the full document input $X$ and the masked input $X \odot M$, resulting in a representation-efficient mask.
Contrary to the belief that estimating mutual information is inherently challenging~\cite{poole2019variational, belghazi2018mine}, recent studies~\cite{skean2024frossl, yu2019multivariate} have demonstrated that it can be computed deterministically when combined with kernel density estimation~\cite{yu2019multivariate, giraldo2014measures}. 
This approach to quantifying mutual information is deterministic and integrates seamlessly with gradient descent over mini-batches. 
However, despite extensive experimentation, this approach did not yield the expected results. A potential limitation may stem from the model's sensitivity to hyperparameter tuning, particularly when dealing with high-dimensional image data, which could have restricted its capacity to reach optimal performance.

As an alternative, we propose a novel step by integrating a continuity loss, as known as anisotropic total variation~\cite{bui2023weighted, johnson2016perceptual} combined with the L1 norm to enforce sparsity and smoothness in the generated relevance masks. The L1 norm promotes sparsity by penalizing the magnitude of mask values, ensuring that only the most essential regions of the input are highlighted, thereby producing concise and focused explanations. In parallel, the continuity loss enhances spatial coherence by minimizing abrupt transitions between adjacent regions in the mask. This is achieved by penalizing the L1 norm of horizontal and vertical differences in the reconstructed 2D relevance mask. Together, these complementary objectives enable the model to highlight minimal yet structurally coherent regions of importance. By fostering smoothness and minimality, this approach mitigates the risk of producing noisy or fragmented masks, ensuring that the explanations are not only interpretable but also intuitively aligned with human understanding.

\subsection{Learning context-aware explanations through token interactions}
\label{sec:colpali}
Through our experiments with training the model using the objective function in~\cref{eq:IB_basic}, we identified that the learned mask often overfits to regions directly corresponding to the answer (see \cref{fig:interactivity1} in Appendix). This overfitting reduces the mask's utility, rendering it a mere visual reproduction of the textual answer rather than providing meaningful, context-aware insights. Such masks lack the ability to highlight broader, equally relevant information related to the question, leading to unreliable explanations.

To address this issue, we emphasize the need to regularize the mask-learning process, not only to mitigate overfitting but also to enable learning more generalizable explanations. 
Specifically, we aim to create context-aware explanations that identify and integrate the most relevant information across the broader input space.
To achieve this, we incorporate a publicly available, pretrained vision-language understanding model as a source of prior knowledge to guide mask learning. 
We utilize its inference results of this model as prior information to facilitate interactive mask learning within our framework.
While many models can be used for this purpose, we employ ColPali~\cite{faysse2024colpali} in our implementation. 
ColPali is a vision-language retrieval model. Its architecture integrates visual patches and textual tokens to generate  multi-vector representations. These representations are then subjected to a late interaction matching mechanism~\cite{liu2024analysis, khattab2020colbert}, which maximizes the interaction between each question token and its corresponding relevant tokens in the document image. This provides a degree of explainability by visualizing question-document interactions through heatmaps, offering insights into the model's decision-making process.  Although these visualizations often fail to align with human intuition or provide comprehensive explanations due to their reliance on attention distributions, we use them as a prior mask and refine them within our \MET framework to enhance relevance and interpretability.





\subsection{Proposed learning objective}
Our model achieves the following: (1) it learns to make predictions with masked input documents via cross-entropy loss $\mathcal{L}_{CE}$ (Sec.\ref{sec:ce}), (2) it minimizes the highlighted regions to ensure concise and focused explanations via $\mathcal{L}_{L1}$ (Sec.\ref{sec:l1}), and (3) it generates more context-aware explanations through tokens interaction with the ColPali signals via $ \mathcal{L}_{MSE}$ (Sec.\ref{sec:colpali}). The complete objective function $\mathcal{L}_\textit{VXQA}$ is defined as: 
\begin{equation}
\begin{aligned}
\min \mathcal{L}_\textit{VXQA} =& \gamma \mathcal{L}_{MSE}(M_{p};M)\\ 
&+ \beta\mathcal{L}_{L1}(X; X \odot M) + \mathcal{L}_{CE}(Y;\hat{Y}),\\  
\end{aligned}
\end{equation}  
where \(\mathcal{L}_{MSE}\) aligns the learned mask with the prior $M_p$ derived from ColPali, \(\mathcal{L}_{L1}\) encourages sparsity in the highlighted regions, and \(\mathcal{L}_{CE}\) represents the cross-entropy loss for prediction accuracy. The hyperparameters \(\gamma\) and \(\beta\) balance the contributions of each term.
All components of the objective function are optimized simultaneously in an end-to-end fashion. As explanations $M$ emerge naturally through this integrated optimization process, our framework inherently exhibits \emph{self-explainability}.

\subsection{Postprocessing}
For a more human understandable relevance maps, a postprocessing stage is applied, consists in enclosing the connected relevance regions with bounding boxes and keep only $k$ boxes with high relevance score. This step, applied only in the inference stage, is further detailed in \cref{subsec:postprocessing}.



\section{Experiments and Results}
\label{sec:experiments}
\begin{table*}[t]
\vskip 0.15in
\begin{center}
\begin{small}
\begin{sc}
\resizebox{\linewidth}{!}{
\begin{tabular}{cccccccc}
\toprule
\textbf{Method} & \textbf{Mask Thresh.} 
& \multicolumn{3}{c}{\textbf{DocVQA}~\cite{mathew2021docvqa}} 
& \multicolumn{3}{c}{\textbf{{PFL-DocVQA}}~\cite{tito2024privacy}} \\
\cmidrule(lr){3-5} \cmidrule(lr){6-8}
 &  
 & Acc. $\uparrow$ & ANLS $\uparrow$ & Pix. Ratio $\downarrow$
 & Acc. $\uparrow$ & ANLS $\uparrow$ & Pix. Ratio $\downarrow$ \\
\midrule
\makecell{Pix2Struct, unmasked \\ \cite{lee2023pix2struct}} &	-- &	0.56	&0.68	&1 &0.80	&0.92	&1\\
Our model, unmasked &	--	&0.51	&0.65	&1 & 0.57	&0.79	&1\\

\addlinespace
\midrule

\multirow{4}{*}{Raw Attention} & 0.05	&0.34	&0.46	&0.38 &0.06	&0.13	&0.27\\
    &0.10	&0.20	&0.32	&0.20 &0.04	&0.10	&0.17\\
    &0.25	&0.07	&0.13	&0.08 &0.01	&0.04	&0.09 \\
    &0.50	&0.03	&0.08	&0.04 & 0.00	&0.01	&0.05 \\
\midrule
\multirow{2}{*}{\makecell{Attention Rollout\\\cite{abnar2020quantifying}}} & 0.01	&0.03	&0.07	&0.04 &0.00	&0.02&	0.03\\
& $1 \times 10^{-5}$	&0.06	&0.11	&0.07 &0.00	&0.03	&0.05\\
\midrule
\makecell{Grad-CAM \\\cite{selvaraju2017grad}} & 0.50	&0.01	&0.05	&0.18 & 0.00	&0.03	&0.16 \\ 
\midrule
\makecell{ColPali$+$Pix2Struct\\ \cite{faysse2024colpali}} & 0.50	&0.38	&0.50	&0.23 & 0.28	&0.39	&0.18 \\
\midrule
\multirow{2}{*}{\textbf{Ours}} & 0.70	&\textbf{0.38}	&\textbf{0.54}	&\textbf{0.23} & \textbf{0.43}	&\textbf{0.66}	&\textbf{0.22} \\
&0.80	&0.30	&0.46	&0.11 & 0.33	&0.54	&0.13\\
\bottomrule
\end{tabular}}
\end{sc}
\end{small}
\end{center}
\vskip -0.1in
\caption{Evaluation of various explainability methods under different mask threshold settings on the DocVQA task. The results demonstrate the trade-offs between the interpretability and utility of the different approaches. The unmasked approaches are set for reference as upper bounds for utility.}
\label{tab:sota}
\end{table*}

\subsection{Experimental Setup}
We posit that a good explanation map should highlight all the critical regions necessary to answer a question while minimizing irrelevant areas correctly, thus being both sufficient and minimal. We experimentally compare our method against several baseline techniques to evaluate both qualities. The evaluation process is structured as follows: explanation masks are generated using different methods, then postprocessed and binarized to keep only the relevant information when applied to the original document image. After that, masks are applied and the resulting image is passed to our fine-tuned Pix2Struct to predict the answer. Predicted answer is compared with the ground truth (GT) using accuracy and average normalized Levenshtein similarity (ANLS) to assess the sufficiency. Furthermore, the proportion of highlighted relevant regions in the mask relative to the total image area, referred to as the pixel ratio, is calculated to measure mask minimality. An ideal method 
achieves a favorable trade-off between explanation utility (accuracy and ANLS) and minimality (pixel ratio). The experiments are done on two datasets, DocVQA~\cite{mathew2021docvqa} and PFL-DocVQA~\cite{tito2024privacy}.
\subsection{Baselines}
We compare our approach against three categories of baseline methods:
\begin{itemize}[leftmargin=*]
    \item Attention Baselines: In this category, we include both raw attention (obtained from Pix2Struct's last layer) and attention rollout \cite{abnar2020quantifying}, which use aggregated attention weights across all layers to provide a more holistic view of the model’s focus.
     \item Gradient Baselines: We implemented Grad-CAM \cite{selvaraju2017grad}, adapting it for the autoregressive sequence output layer by applying it to each output token from the decoder and aggregating the resulting maps.

    \item Retrieval Baseline: This approach utilizes ColPali \cite{faysse2024colpali}, a retrieval-based framework. Here, the input (question + answer) is first processed by ColPali on its own to generate a heatmap as the explanation. This heatmap is then applied to the image, which is subsequently passed to the Pix2Struct model for prediction. We call this baseline ColPali$+$Pix2Struct.
\end{itemize}

\subsection{Results}
\paragraph{Quantitative Evaluation}
The results obtained are summarized in \cref{tab:sota}, where different methods are evaluated to balance utility and explainability. As a reference for upper bound utility, we include results for unmasked images using the original Pix2Struct model and our trained model. Although our model exhibits a slight decrease in performance, this can be attributed to the challenge of simultaneously managing both masked and clear images during training. 

\begin{figure*}[t]
\vskip 0.2in
\begin{center}
\begin{tabular}{|c|c|c|c|}
 \toprule
  \includegraphics[width=0.22\linewidth]{icml2025/imgs/236/original.png}   & 
   \includegraphics[width=0.22\linewidth]{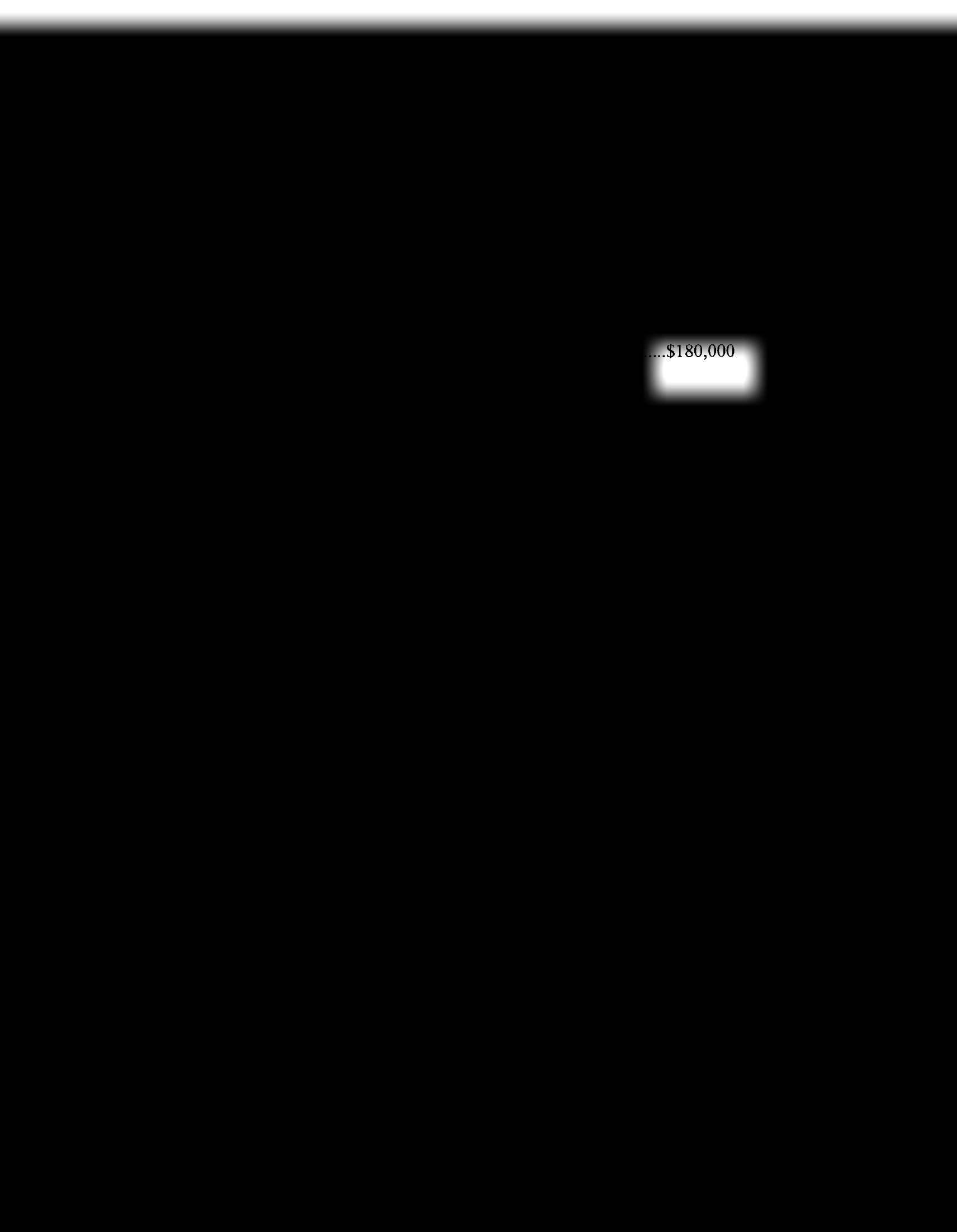}   & 
    \includegraphics[width=0.22\linewidth]{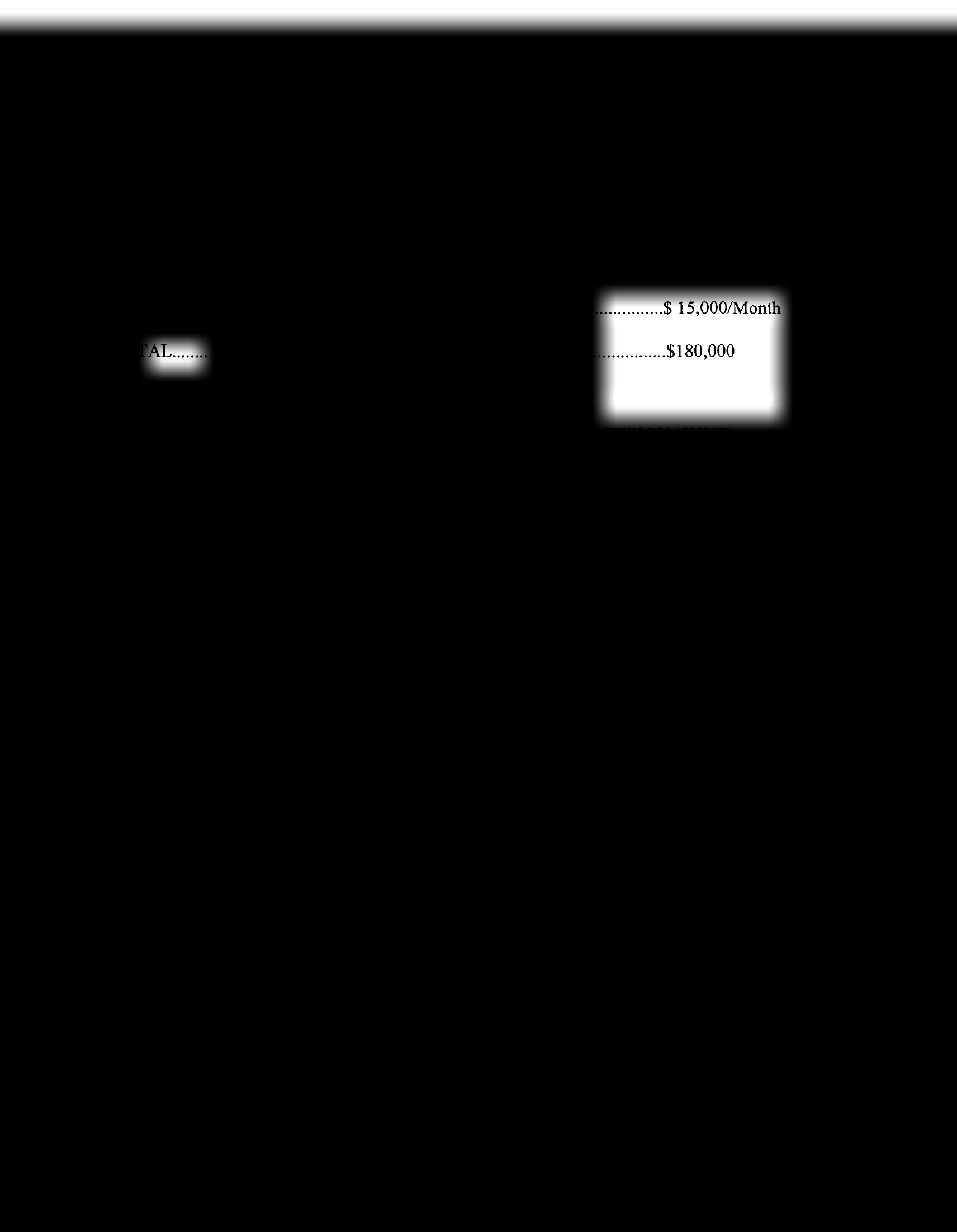}  & 
    \includegraphics[width=0.22\linewidth]{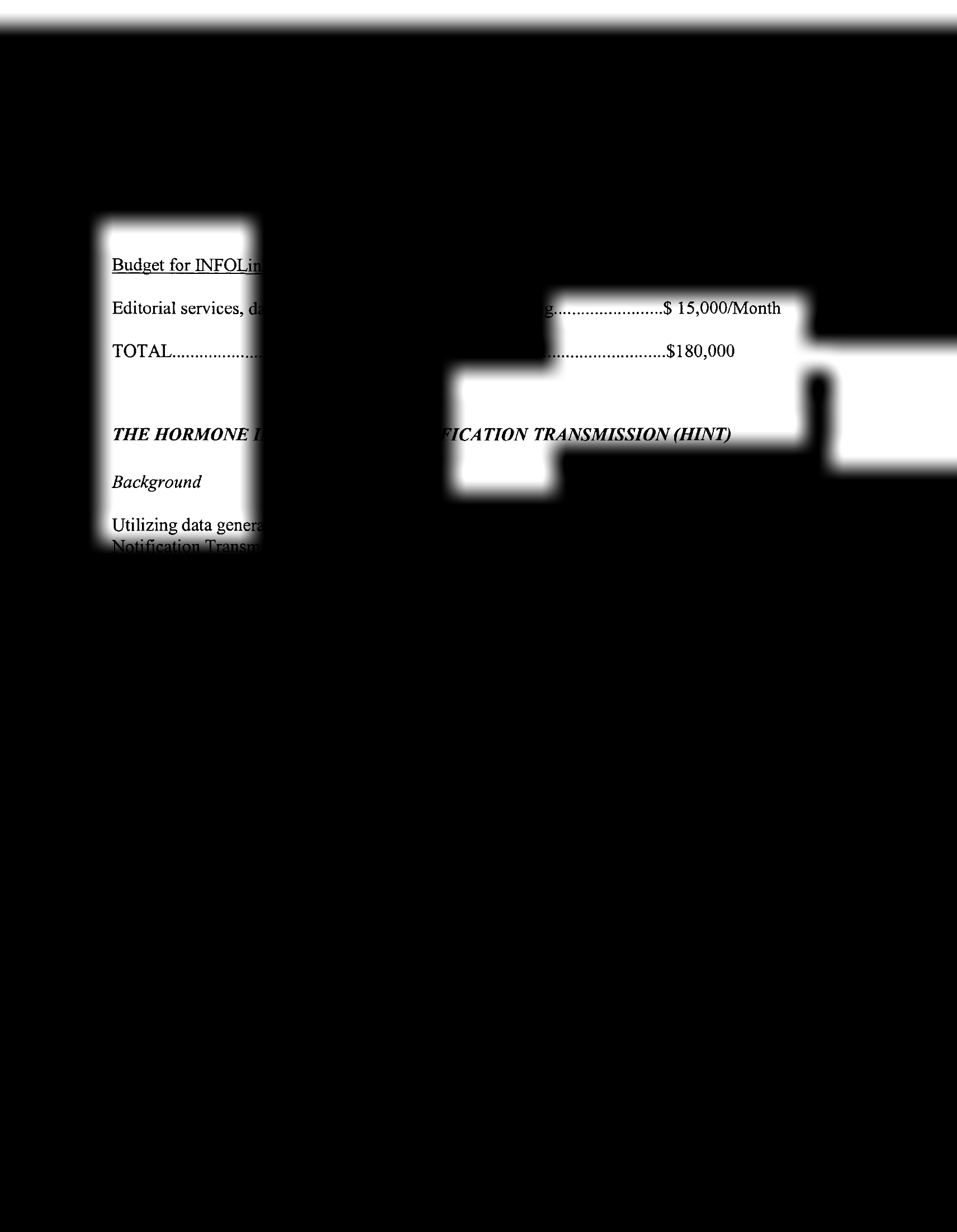}  \\ \\ Original Image & Raw Attention (0.5)&  Raw Attention (0.1)   &  Raw Attention (0.05) \\
    \midrule
     \includegraphics[width=0.22\linewidth]{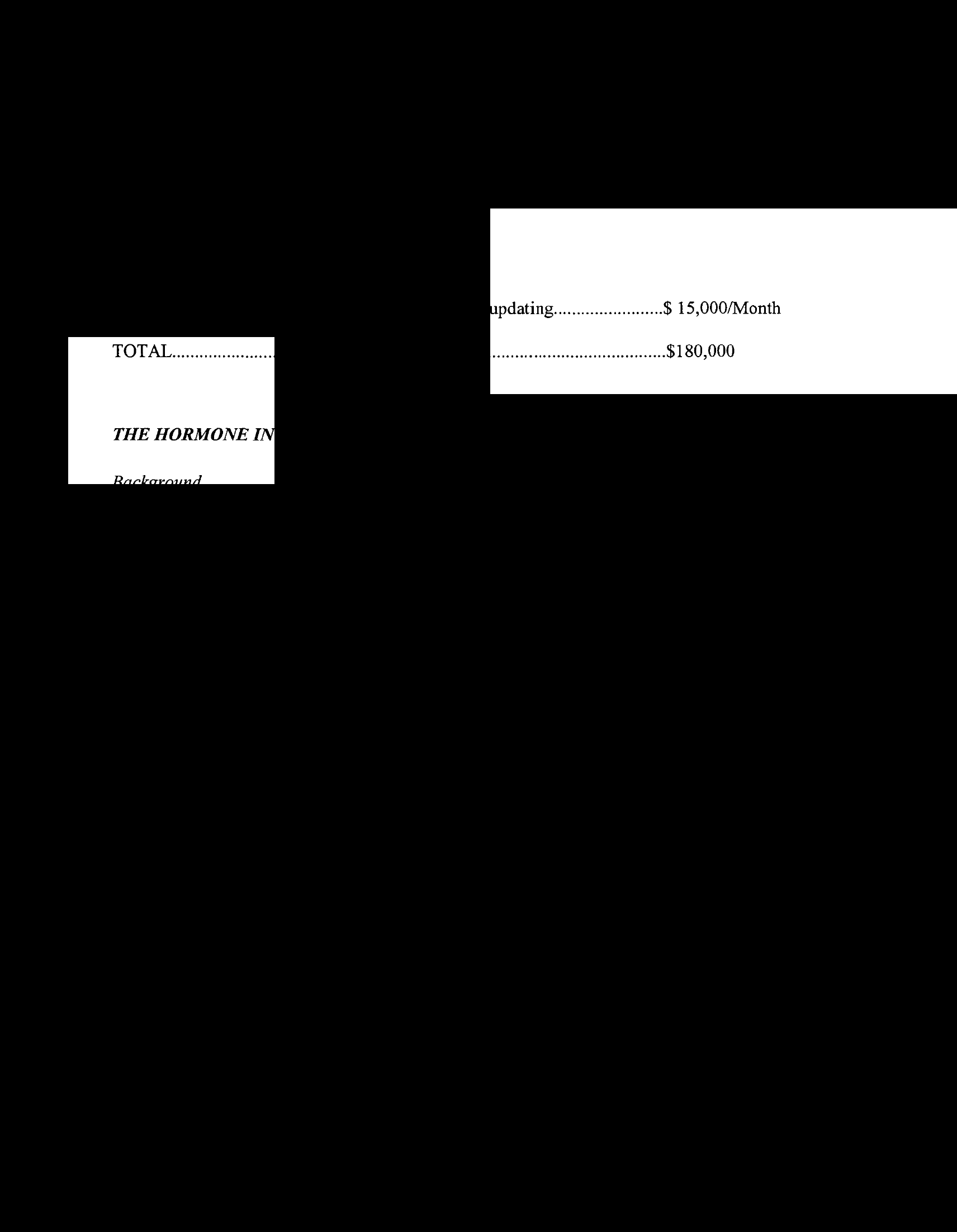}  & 
      \includegraphics[width=0.22\linewidth]{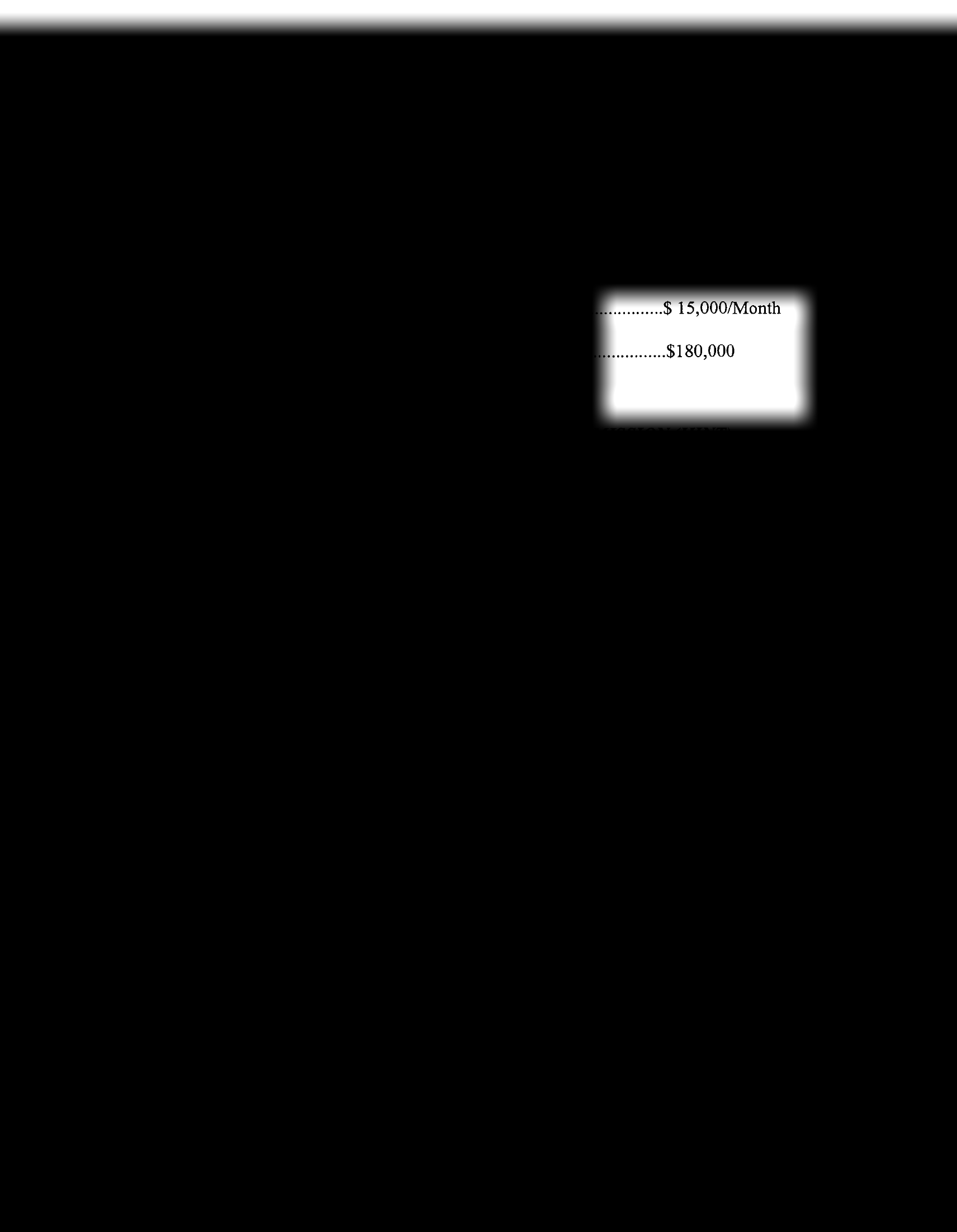}
      &
     \includegraphics[width=0.22\linewidth]{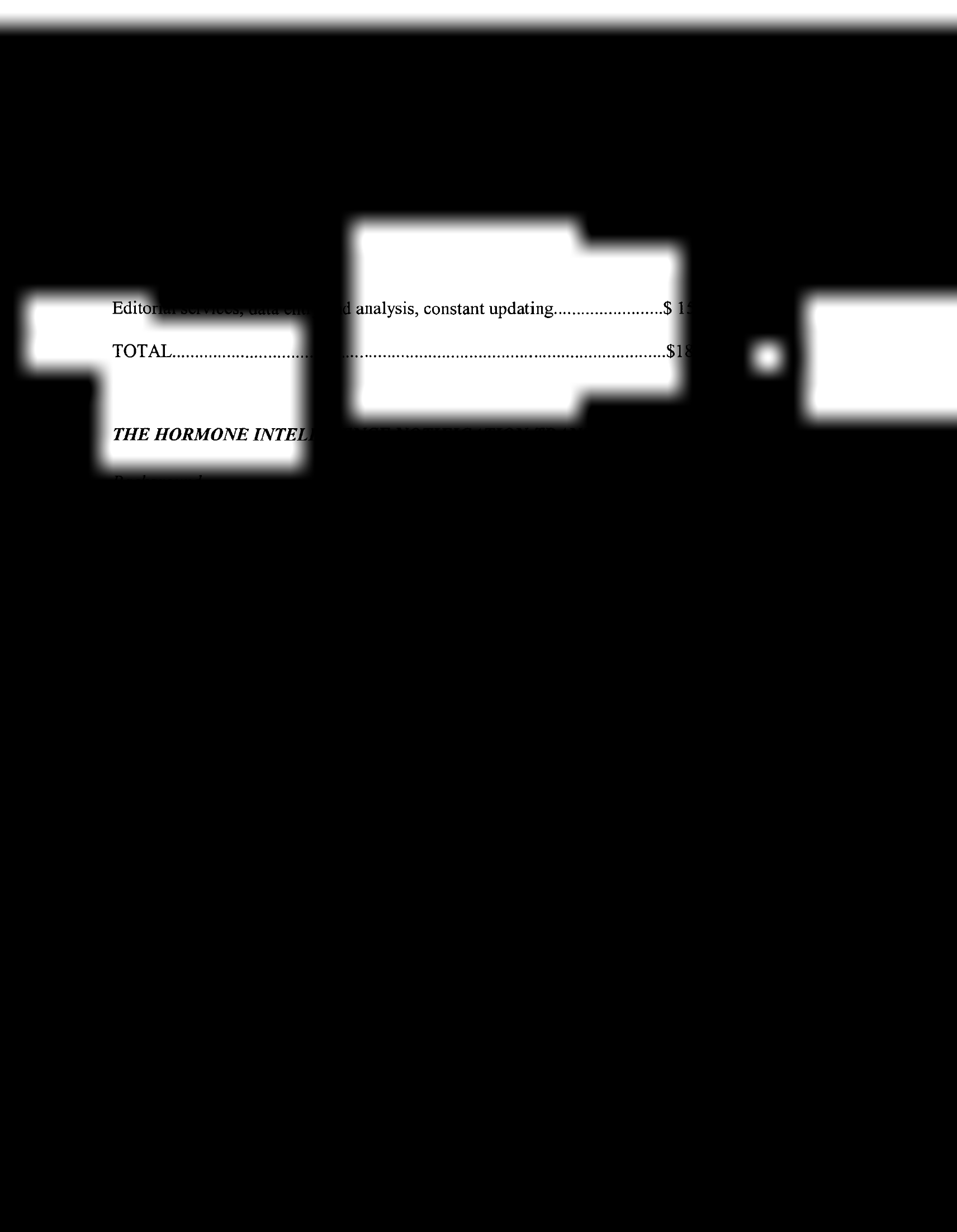} 
     &
     \includegraphics[width=0.22\linewidth]{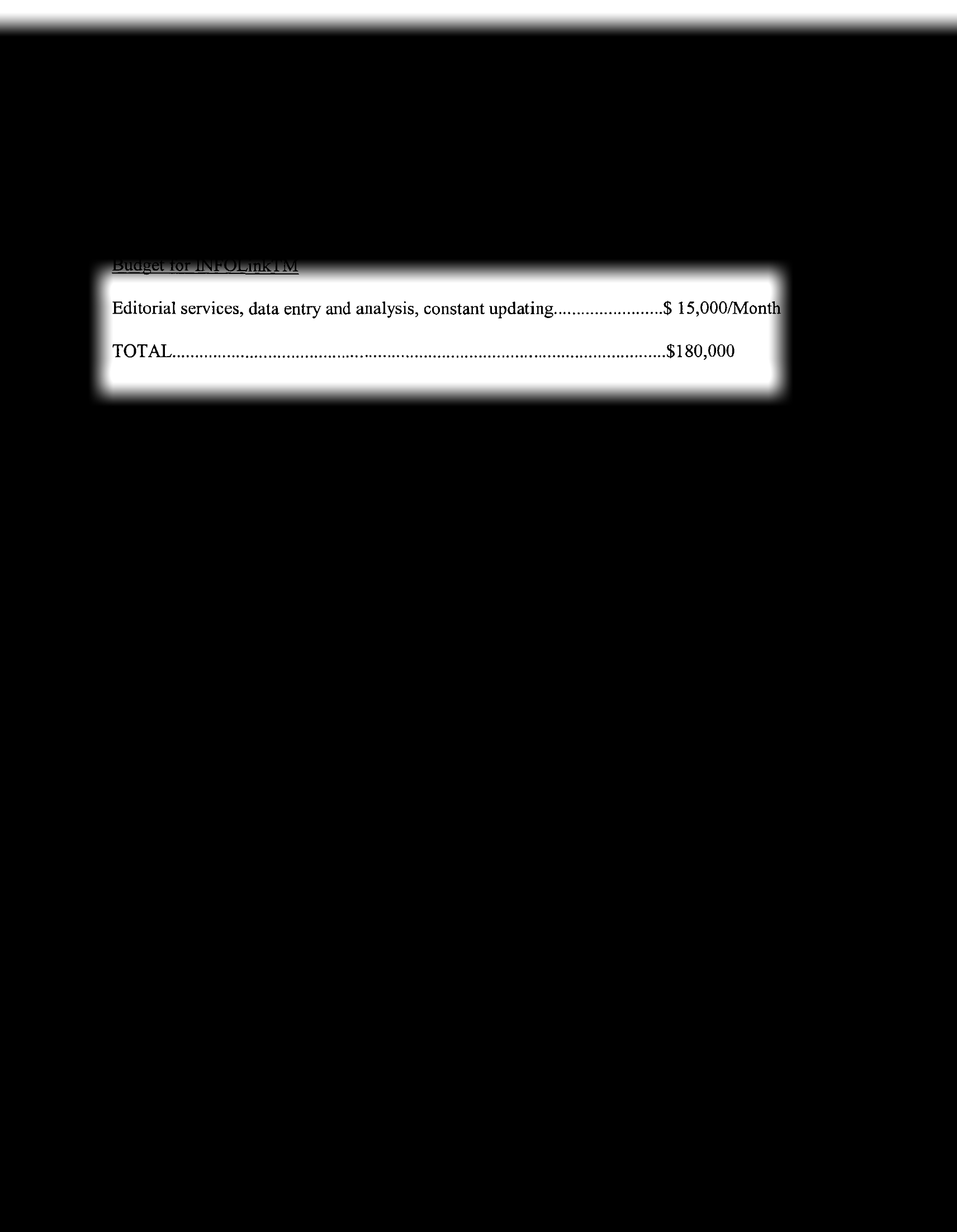}
\\ ColPali$+$Pix2Struct & Att. rollout & Grad-CAM  & \textbf{Ours (0.7)} \\ 
\bottomrule 
\end{tabular}
\caption{A comparison of explanations generated by different methods for the question \textbf{``What is the total amount?''} with the model’s answer being \textbf{``\$180,000''}. Relevance maps with different (thresholds) are applied to the input image to keep  only the relevant regions.
Best viewed at high zoom. Additional qualitative results across diverse contextual scenarios are provided in~\cref{sec:examples}.}
\label{fig:qualitative_results}
\end{center}
\vskip -0.2in
\end{figure*}

Our approach stands out among the explainable methods, achieving the best utility in terms of accuracy and ANLS while maintaining a modest pixel ratio. This highlights the method’s ability to focus on the most critical regions of the input image, providing concise yet effective explanations. In contrast, raw attention exhibits reasonable performance when the mask threshold is set very low, as this results in large pixel ratios that cover a significant portion of the image. However, as the threshold increases, raw attention's utility metrics degrade significantly, suggesting it struggles to maintain performance when forced to operate with smaller, more answer-overfitted explanations.
Attention rollout, while offering a more comprehensive aggregation of attention weights across layers, performs poorly overall. Its failure to accurately localize relevant regions is likely due to the propagation of errors (missing the contextual regions) across layers, which focused the relevant regions only around the answer.  Grad-CAM, despite its popularity, also shows weak performance in this context. This can be attributed to the challenges of adapting Grad-CAM to autoregressive sequence models, where token-level aggregation may reduce its ability to generate precise explanations. Finally, the ColPali$+$Pix2Struct approach demonstrates competitive utility metrics with reasonable pixel ratios on the DocVQA dataset, however, our method achieves even better utility at the same ratio, as well as significantly better performance on the PFL-DocVQA dataset, which further shows the efficacy of our approach.

\paragraph{Qualitative Evaluation}
To reflect the results presented in \cref{tab:sota}, we illustrate the relevance masks generated by different methods in \cref{fig:qualitative_results}. Each method was tasked with explaining the model’s response to the question \emph{``What is the total amount?''}, where the predicted answer is \emph{``\$180,000''}. The explanation masks are overlaid on the input image to highlight regions deemed relevant for the prediction. 
From the visualizations, it is evident that our method produces explanations that are both precise and comprehensive. Compared to baseline methods, such as ColPali$+$Pix2Struct, Attention Rollout, and Raw Attention, our approach successfully isolates all the textual elements contributing to the final answer in a relevance map that is easy to visually process by a human, while avoiding irrelevant regions. Notably, Grad-CAM focuses on certain parts of the document that do not contain all the required information to answer the question by the model.
\begin{figure}[t]
\vskip 0.2in
\begin{center}
\includegraphics[width=0.99\linewidth]{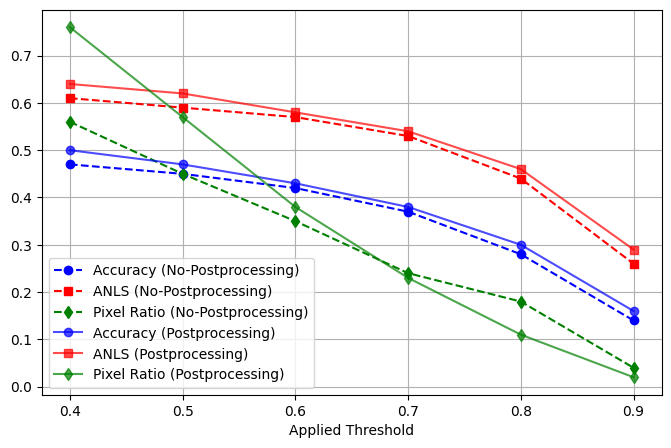}
\caption{Performance of our method with and without postprocessing, under different thresholds applied to the relevance masks.}
\label{fig:res_postprocessing}
\end{center}
\vskip -0.2in
\end{figure}

When these masked images were fed back into the model, all methods except Grad-CAM produced the correct answer. However, this does not necessarily indicate good relevance maps. In such scenarios, highlighting only any number will lead the DocVQA to use it as an answer, even without evidence that it corresponds to the ``total''. 
Our method leverages the most clear and \emph{context-aware} explanation. This results in an accurate visualization of the reasoning process, which enhances users' ability to understand the model’s decision-making, and fosters greater trust and confidence in the system.

\paragraph{Effect of Post-processing}
To study the impact of postprocessing and thresholding on the resultant masks, we perform the experiment presented in \cref{fig:res_postprocessing}. The experiment highlight the trade-offs between precision and coverage when applying different thresholds and when using postprocessing. Lower thresholds achieve higher Accuracy and ANLS but at the cost of significantly larger Pixel Ratios, indicating less precise localization of relevance regions. As the threshold increases, Pixel Ratio decreases, leading to more focused masks but with a corresponding decline in Accuracy and ANLS. When postprocessing is applied, the results show an improvement in the accuracy and ANLS with a reduction in Pixel ratio with a threshold $\geq$ 0.7. This improvement underscores the importance of postprocessing as a critical step in refining the outputs of our method, balancing relevance and mask quality.

\paragraph{Learning Objectives} 
In our method, three learning objectives are used during training: Sufficiency (S), Minimality (M), and Token Interactions (TI). To evaluate their individual and combined contributions to performance, we conducted an ablation study, as shown in \cref{tab:ablation_objectives}. The first row, where only the sufficiency loss is applied, achieves the highest Accuracy and ANLS but has the largest Pixel Ratio. This indicates that the model identifies the entire image as relevant, maximizing coverage but lacking precision. When both sufficiency and minimality losses are employed, the Pixel Ratio drastically decreases to $0.02$, demonstrating the effectiveness of minimality in constraining the mask to small regions that are necessary to answer the question. However, this extreme reduction leads to overfitting, as the model disregards accompanying context that may be important for generalization. Consequently, both Accuracy and ANLS significantly decrease on the test data, highlighting a trade-off between precision and informativeness. Finally, the inclusion of token interactions loss (S+M+TI) balances these trade-offs, improving mask coverage on the context leading to achieving significant gains in Accuracy and ANLS. 
\begin{table}[t!]
\vskip 0.15in
\begin{center}
\begin{small}
\begin{sc}
\begin{tabular}{cccccc}
\toprule
\textbf{S} & \textbf{M} & \textbf{TI} & \textbf{Acc. $\uparrow$} & \textbf{ANLS $\uparrow$} & \textbf{Pixel Ratio $\downarrow$} \\
\midrule
\checkmark & & &0.51	&0.65	&1 \\
\checkmark &\checkmark & &0.19	&0.36	&0.02\\
\checkmark &\checkmark &\checkmark & 0.38	&0.54	&0.23\\
\bottomrule
\end{tabular}
\end{sc}
\end{small}
\end{center}
\vskip -0.1in
\caption{Ablation study on the impact of different learning objectives. S: sufficiency, M: minimality, TI: Token Interactions.}
\label{tab:ablation_objectives}
\end{table}

\begin{table}[t!]
\vskip 0.15in
\begin{center}
\begin{small}
\begin{sc}
\resizebox{\columnwidth}{!}{
\begin{tabular}{lcc}
\toprule
\textbf{Method} & \textbf{Context $\uparrow$} &  \textbf{Clarity $\uparrow$}\\
\midrule
Raw Attention (0.25) & 2.90 $\pm$ 0.48 & 2.75 $\pm$  0.42\\
\midrule
Raw Attention (0.50) & 2.26 $\pm$ 0.54& 2.34 $\pm$ 0.60 \\
\midrule
ColPali$+$Pix2Struct & 3.97 $\pm$ 0.25& 3.02 $\pm$ 0.42\\
\midrule
\textbf{Ours (0.7)} & \textbf{4.49} $\pm$ \textbf{0.26} & \textbf{3.56} $\pm$ \textbf{0.41}\\
\bottomrule
\end{tabular}}
\end{sc}
\end{small}
\end{center}
\vskip -0.1in
\caption{Results of the human evaluation of the explanation quality for different methods. Participants rated on a scale of 1 to 5.}
\label{tab:human evaluation}
\end{table}
\paragraph{Human Evaluation}
The goal of developing explainable DocVQA models is to enhance human trust in these systems. To experimentally assess the quality of the explanations, we conducted a human evaluation study with $42$ impartial participants, blind to the details of our method. We used $10$ example document images. For each image, participants were shown a corresponding question, the ground truth (GT) answer, and four different explanation masks generated from relevance maps of the top-performing methods seen in \cref{tab:sota}, based on the trade-offs between Accuracy, ANLS, and Pixel Ratio--this includes ColPali, raw attention, and our own method. Participants were asked to evaluate each explanation on two key criteria, rated on a scale of 1 to 5:

\begin{itemize}[leftmargin=*]
    \item \textbf{Context-awareness of the explanation:} Participants responded to the question: ``How confident are you  (1-5) to answer the question using only this masked image?'' This question assesses whether the explanation directly supports the answer and provides all the necessary context to answer.
    \item \textbf{Clarity of the explanation:} Participants responded to the question: ``Does the masked image include only all the necessary information in a clear, concise, and comprehensive way to fully explain the model’s reasoning? Rate on a scale (1-5).'' This question evaluates the extent to which the explanation provides thorough, focused and human comprehensive coverage of the relevant content.
\end{itemize}
The results of the human evaluation are presented in \cref{tab:human evaluation}. As shown, our method outperforms all baselines in both context-awareness and clarity, achieving the highest scores, compared to ColPali and Raw Attention variants. These results highlight our method’s ability to generate clear explanations that enable participants to confidently answer questions without the original image, while also providing comprehensive and focused masks that include only the necessary information. Thus, using our method enhances user trust and understanding of the model's predictions.

Furthermore, we conducted another human preference study to directly compare the masks of our method against the top-performing baseline in the previous study, that is ColPali+Pix2Struct. In total, $12$ participants evaluated $21$ randomly selected question-answer pairs ($252$ trials in total). As a result, our method was preferred in $163$ trials ($64.7\%$; $95\%$ CI [$58.4\%$, $70.6\%$], $p << 0.001$), with all participants ($12/12$) favoring our approach overall. Thus, our method is offering a significantly more compact and interpretable explanations than ColPali+Pix2Struct.

\paragraph{Model Agnosticity}
Our \MET framework is designed to be model-agnostic. To demonstrate its architectural flexibility, we implemented it with Donut~\cite{kim2021donut} as an alternative to Pix2Struct. Notably, Donut and Pix2Struct differ fundamentally in how they incorporate the question. Pix2Struct renders the question directly onto the input image, while Donut tokenizes the question and uses it to condition the decoder during generation. To accommodate this difference, we adapt our method by modifying the mask head to also incorporate the encoded question. Qualitative results presented in~\cref{app:sec:donut} illustrate that our method consistently identifies relevant regions across both backbones, validating its model-agnostic design.


\section{Conclusion}
\label{sec:conclustion}
This paper introduced \MET, the first self-explainable model for DocVQA, enhancing interpretability by learning relevance maps that elucidate the model's decisions.
Unlike conventional approaches that rely on post-hoc explanations or merely justify answers without deep contextual grounding, \MET inherently generates visual and context-aware explanations to the answers, ensuring that the model's process remains transparent, interpretable, and aligned with human reasoning.  
This is achieved by formulating and including fundamental explainability principles directly into the learning process. 
Our extensive evaluations demonstrate the effectiveness of our framework.

Future research will explore the generalization of our method across diverse DocVQA architectures and datasets, as well as optimizing the trade-off between explainability and task performance. We envision \MET as a stepping stone toward robust, transparent, and user-centric AI solutions in document intelligence.




\section*{Acknowledgments}
We thank all participants for their contributions to the human preference study.
This work has been supported by the Consolidated Research Group 2021 SGR 01559 from the Research and University Department of the Catalan Government, and by project PID2023-146426NB-100 funded by MCIU/AEI/10.13039/501100011033 and FSE+. 
This work has been funded by the European Lighthouse on Safe and Secure AI (ELSA) from the European Union’s Horizon Europe programme under grant agreement No 101070617. 
The work of Changkyu Choi is supported by Visual Intelligence, a centre for research-based innovation funded by the Research Council of Norway and its consortium partners (RCN grant no. 309439).

% ...

\section*{Impact Statement}
To the best of our knowledge, this is the first work to introduce explainability to the Document Visual Question Answering (DocVQA) domain. As DocVQA systems are increasingly deployed in high-stakes environments such as banking, healthcare, and public administration where automated systems are used to extract information from documents with the goal of minimizing human intervention, the need for transparent and trustworthy models becomes critical. Our work addresses this urgent gap by enabling interpretability of DocVQA models, which is essential for ensuring reliability, building user trust, and facilitating responsible adoption in sensitive real-world applications.

\bibliography{arxiv}
\bibliographystyle{icml2025}

\newpage
\appendix
\onecolumn
\icmltitle{Appendix}

To ensure consistency, we continue the numbering of figures and tables from the main paper. The first references in the Appendix begin with Figure 5 and Table 4.

\section{Implementation Details}
\subsection{Postprocessing}\label{subsec:postprocessing}
\begin{figure}[h]
\vskip 0.2in
\begin{center}
\begin{tabular}{ccc}

  \includegraphics[width=0.30\linewidth]{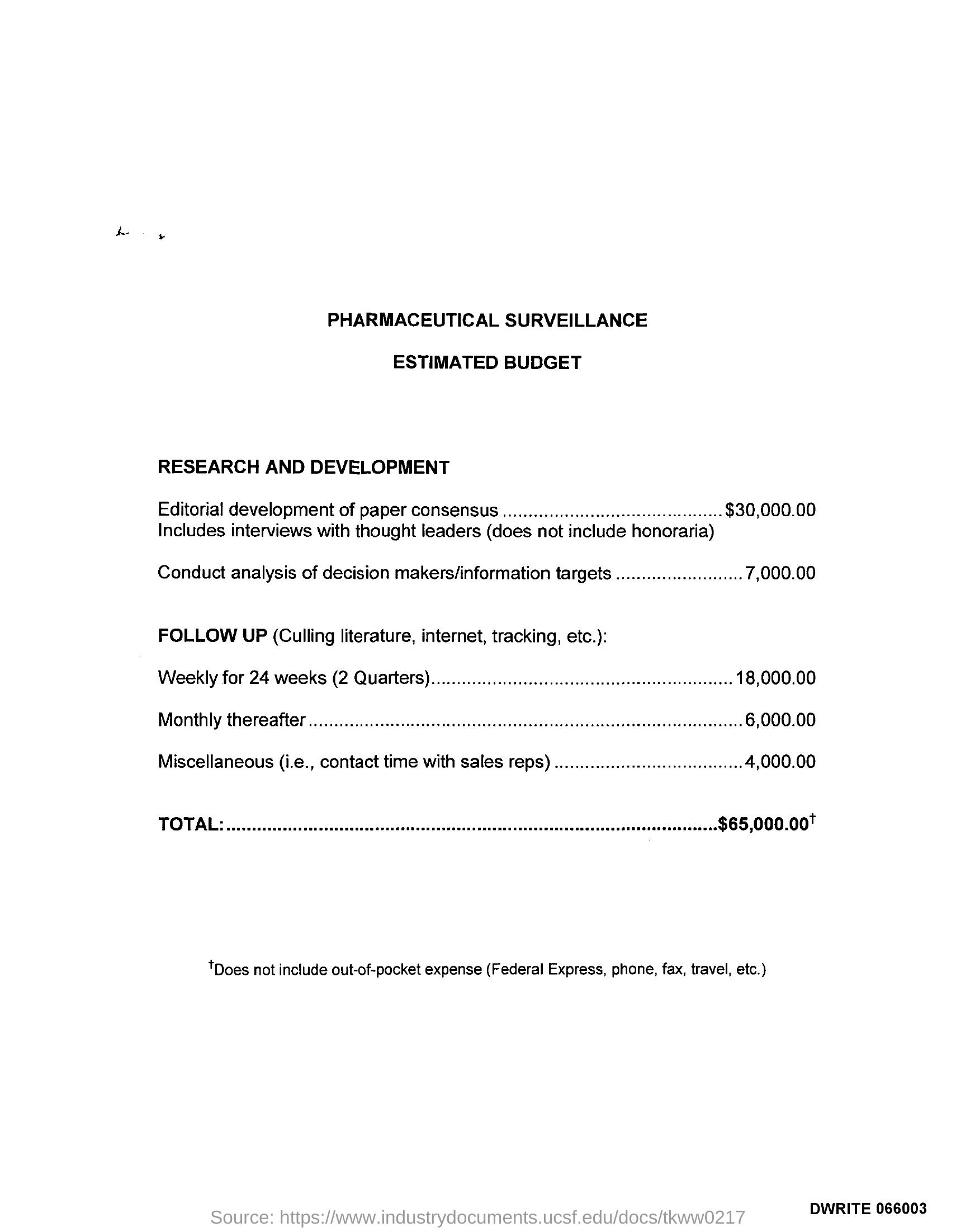} &
  \includegraphics[width=0.30\linewidth]{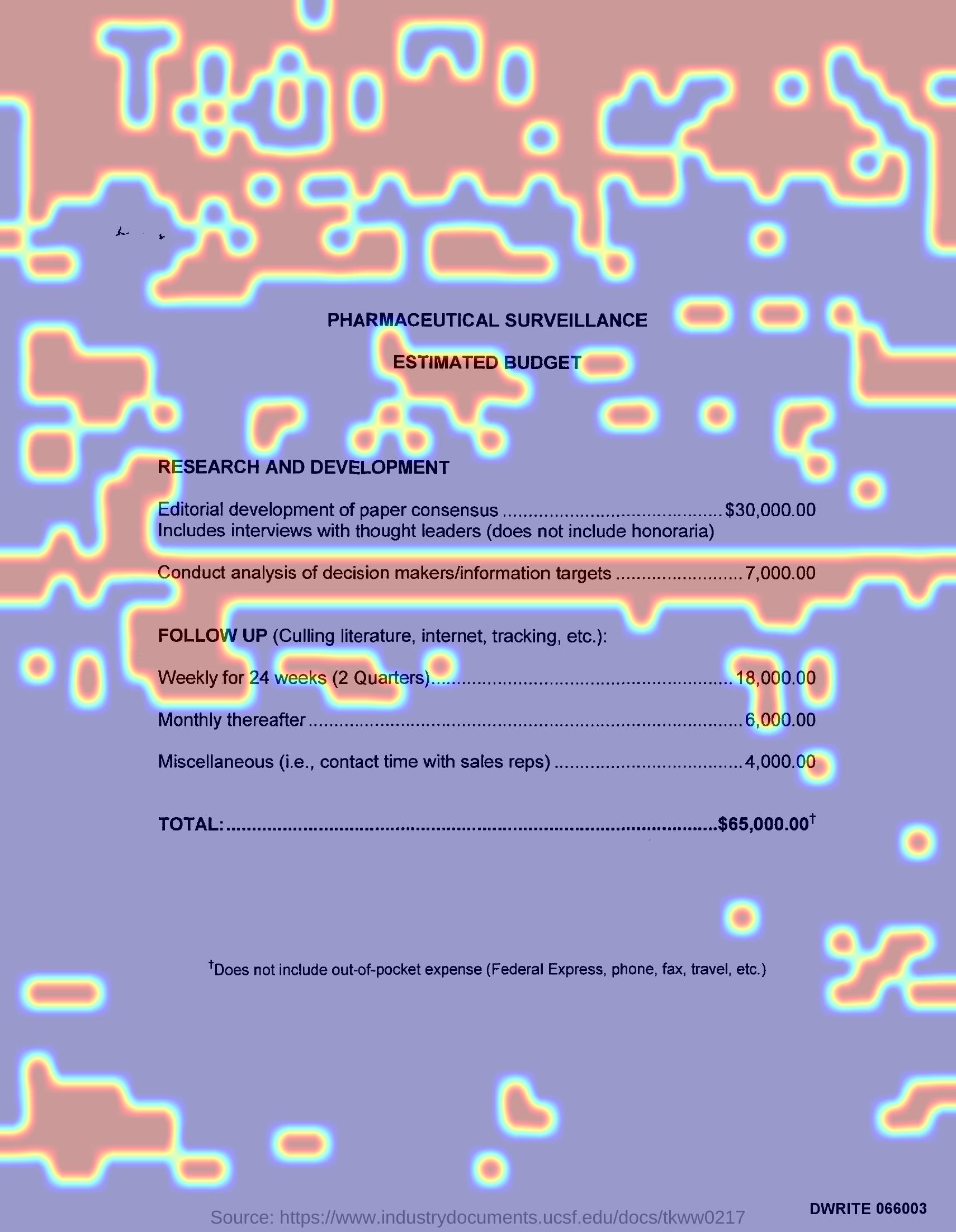} &
  \includegraphics[width=0.30\linewidth]{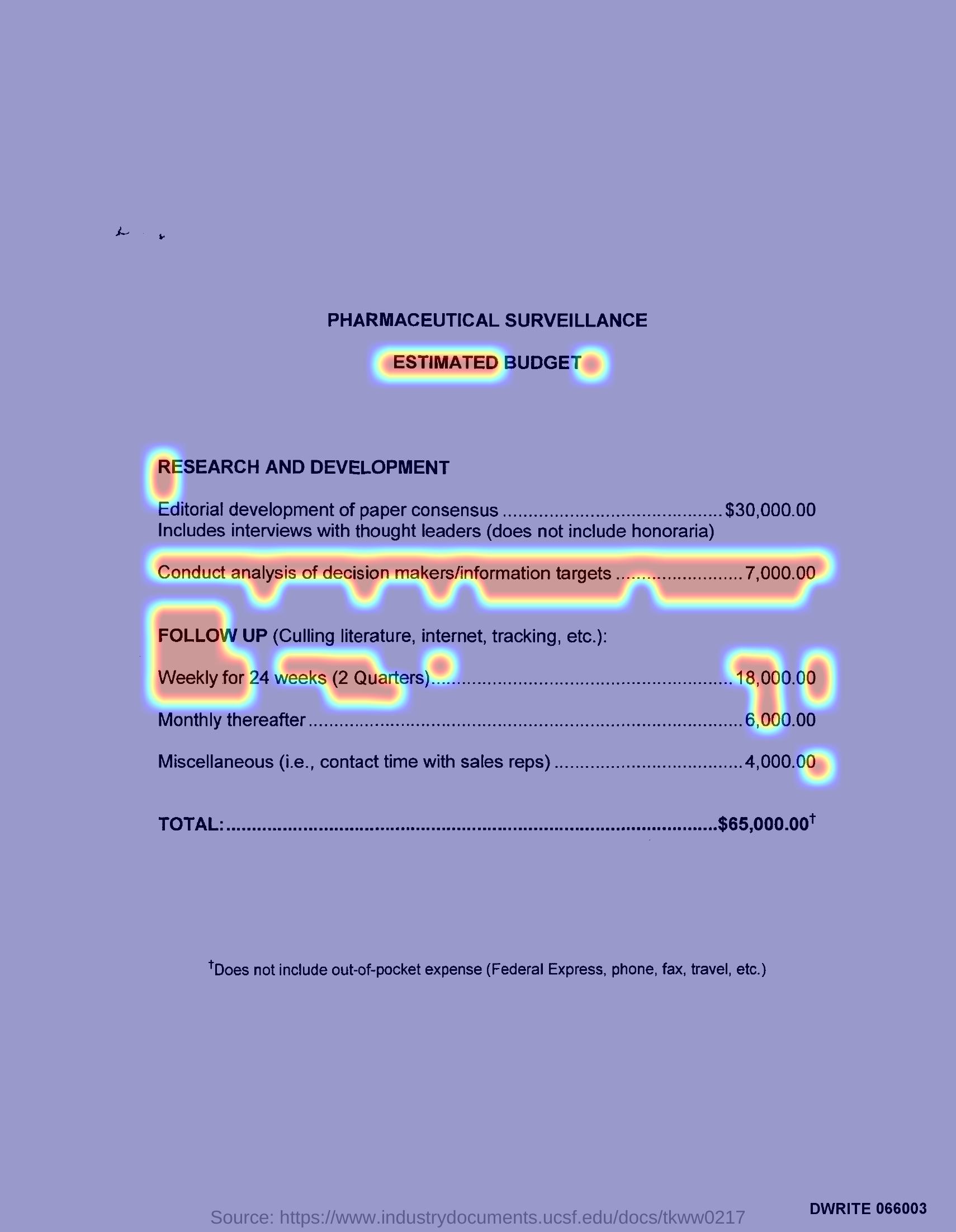} \\
  Original & \makecell{Predicted Mask \\(with bg regions)} & \makecell{Predicted Mask \\(without bg regions)} 
 \\ 
\end{tabular}
\caption{Visualization of the background removal step in our methods. Question: ``What is the estimated budget of ‘conduct analysis of decision makers/ information targets’  in research and development?''. Model answer  ``\$ 7,000.00''.}
\label{fig:bg_rmoval}
\end{center}
\vskip -0.2in
\end{figure}

The relevance map produced by our method is passed throught a postprocessing step that envolve 2 stages: (1) background removal, (2) connected region bounding. In the following we detail each step, noting that for a fair comparison, these three steps are applied to all the methods presented in \cref{tab:sota}:

\paragraph{Background Removal.} Since the output mask may highlight irrelevant background regions, we first filter out the highlighted regions that fall input image tokens that represent background. In other words, we discard tokens whose encoded patch variance are below a threshold of $0.01$, as they are considered background in the document images, where in this domain the background is usually blank. This step ensures that only informative regions contribute to the final relevance map. In \cref{fig:bg_rmoval}, we show an extreme case of this scenario, where the produced relevance map contain a lot of background regions, after applying this postprocessing step, we can see that the quality of the map is significantly improved.

\paragraph{Bounding connected regions.}
After filtering out background regions, the remaining mask effectively highlights relevant information, as shown in \cref{fig:bg_rmoval}. However, the raw mask may still appear fragmented and less intuitive for users. To enhance interpretability, we apply an additional postprocessing step that identifies connected components within the relevance map and encloses them in bounding boxes. We chose this approach because text in documents is best highlighted using bounding boxes, ensuring clear visibility and structured localization. Unlike other forms of heatmaps, which may appear vague or ambiguous, bounding boxes provide a precise, easily interpretable representation of the model's focus which facilitate its processing by human users. 

\begin{figure}[H]
\vskip 0.2in
\begin{center}
\begin{tabular}{ccc}

  \includegraphics[width=0.30\linewidth]{icml2025/imgs/555/555_original.jpg} &
  \includegraphics[width=0.30\linewidth]{icml2025/imgs/555/555_no_bg.jpg}&
  \includegraphics[width=0.30\linewidth]{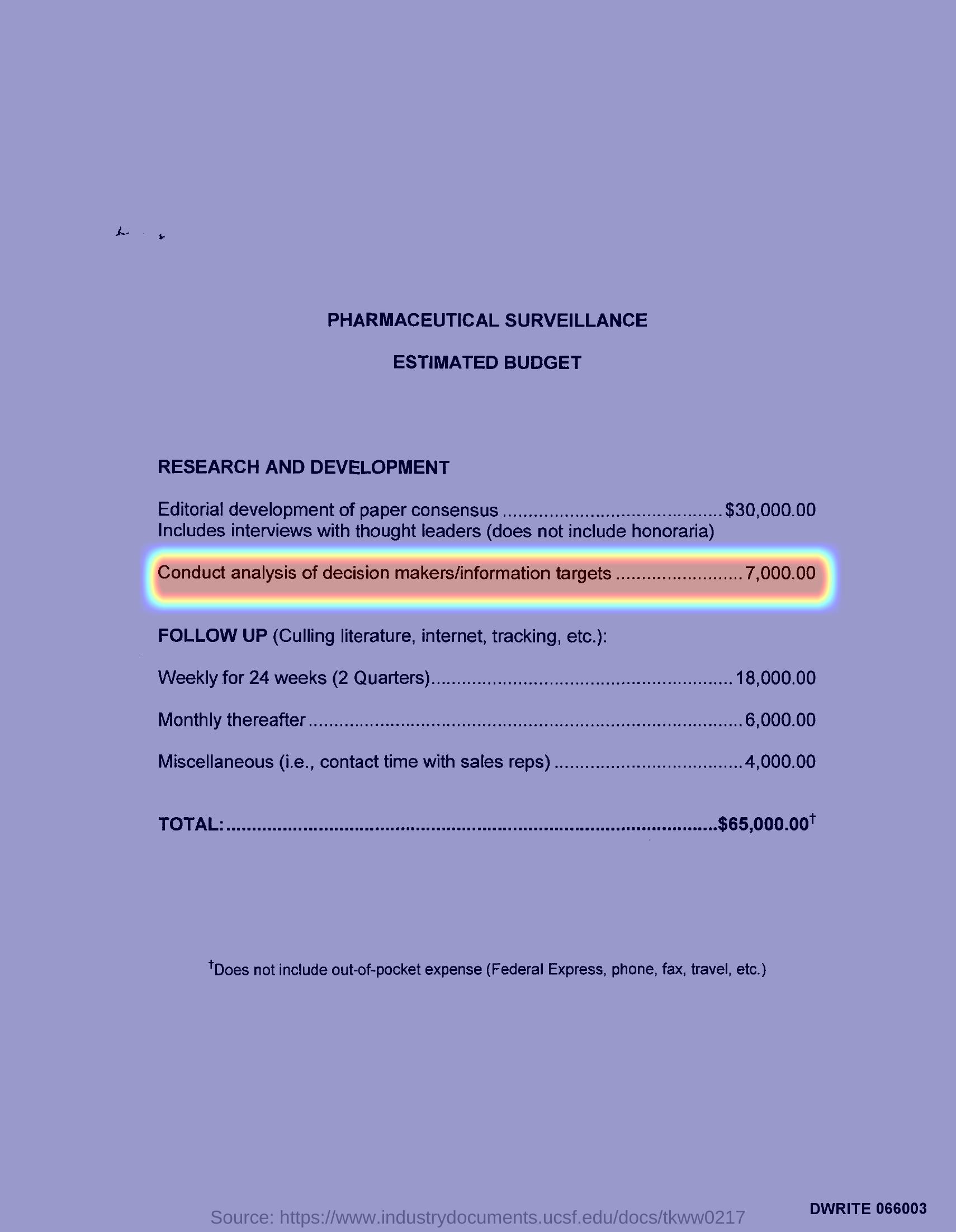} \\
  Original & Predicted Mask & \makecell{Predicted Mask \\(with full postprocessing)} 
 \\ 
\end{tabular}
\caption{Visualization of connecting the relevance regions in the heatmap and keeping only 1 region with top confidence score ($k=1$). Question: ``What is the estimated budget of ‘conduct analysis of decision makers/ information targets’  in research and development?''. Model answer  ``\$ 7,000.00''.}
\label{fig:draw_boxes}
\end{center}
\vskip -0.2in
\end{figure}

After that,  we rank these regions based on their confidence scores and retain only the top $k$ most relevant boxes. This step, shown in \cref{fig:draw_boxes}, ensures that the final explanation is both compact and user-friendly. In our study, we experimented with different values of $k$ and found that $k=3$ provided the best balance between conciseness and completeness. 

\subsection{Mask head}
The implemented mask head is a fully connected neural network (MLP-based decoder) designed to process an input consists of the concatenation of the encoded tokens (features), decoder attentions and the positional encoding. The mask head generates a relevance score for each token, producing a relevance map that highlights the most important regions in terms of influence on the final answer. During training, the output of the mask head is passed through a Sigmoid activation function to ensure that the values are normalized between 0 and 1,  applied to the input image, and then passed to the Pix2Struct component for the masked image prediction.
\subsection{Hyperparameters}

We summarize the values of the hyperparameters used in this paper. We not that the Pix2Struct component is first fine-tuned on the DocVQA dataset~\cite{mathew2021docvqa}, then our proposed model is trained according to the settings presented in \cref{tab:hyperparams}.
Note that in our work, we use the base version of Pix2Struct with 282M parameters. The model is composed of  12 encoder and 12 decoder layers with a hidden size of 768.

\begin{table}[h]
\caption{Hyperparameter Settings}
\label{tab:hyperparams}
\vskip 0.15in
\begin{center}
\begin{small}
\begin{sc}
\begin{tabular}{lc}
\toprule
 \textbf{Hyperparameter} & \textbf{Value} \\
\midrule
Learning Rate & $1 \times 10^{-7}$ \\
Batch Size & 5 \\
Optimizer & AdamW \\
$\gamma$ &  0.5\\
$\beta$ & 5\\

Threshold ($k$) for Postprocessing & 3 \\

\bottomrule
\end{tabular}
\end{sc}
\end{small}
\end{center}
\vskip -0.1in
\end{table}

\section{Additional Results}
\subsection{Selection of  $k$.}
As illustrated in \cref{fig:draw_boxes}, during the post-processing step, we retain only the top-$k$ most relevant boxes. We conducted an ablation study to evaluate the impact of different $k$ values and found that $k=3$ offers the best trade-off between conciseness and completeness. The full results across various $k$ values are reported in \cref{tab:k_selection}. As shown, $k=3$ is considered the best choice for a good trade-off between interpretability and utility.
\begin{table}[h]
\vskip 0.15in
\begin{center}
\begin{small}
\begin{sc}
\begin{tabular}{cccc}
\toprule
\textbf{$k$} & \textbf{Accuracy} & \textbf{ANLS} & \textbf{Pixel Ratio} \\
\midrule
1& 0.36& 0.52& 0.21\\
2                                                      & 0.38                                                     & 0.53                                                      & 0.23                                                              \\
3                                                      & 0.38                                                     & 0.54                                                      & 0.24                                                              \\
4                                                      & 0.39                                                     & 0.55                                                      & 0.25                                                              \\
5                                                      & 0.39                                                     & 0.55                                                      & 0.26                                                              \\
8                                                      & 0.40                                                     & 0.56                                                      & 0.28                                                              \\
10                                                     & 0.41                                                     & 0.57                                                      & 0.29                                                              \\
15                                                     & 0.42                                                     & 0.58                                                      & 0.31                                                              \\
20                                                     & 0.43                                                     & 0.58                                                      & 0.32                                                              \\
50                                                     & 0.45                                                     & 0.59                                                      & 0.38                                                              \\
100                                                    & 0.45                                                     & 0.60                                                      & 0.40     \\                          \bottomrule
\end{tabular}
\end{sc}
\end{small}
\end{center}
\vskip -0.1in
\caption{A study on varying the number of top relevant boxes during post-processing.}
\label{tab:k_selection}
\end{table}

\subsection{The effect of using token interactions.}
\begin{figure}[H]
\vskip 0.2in
\begin{center}
\begin{tabular}{ccc}

  \includegraphics[width=0.30\linewidth]{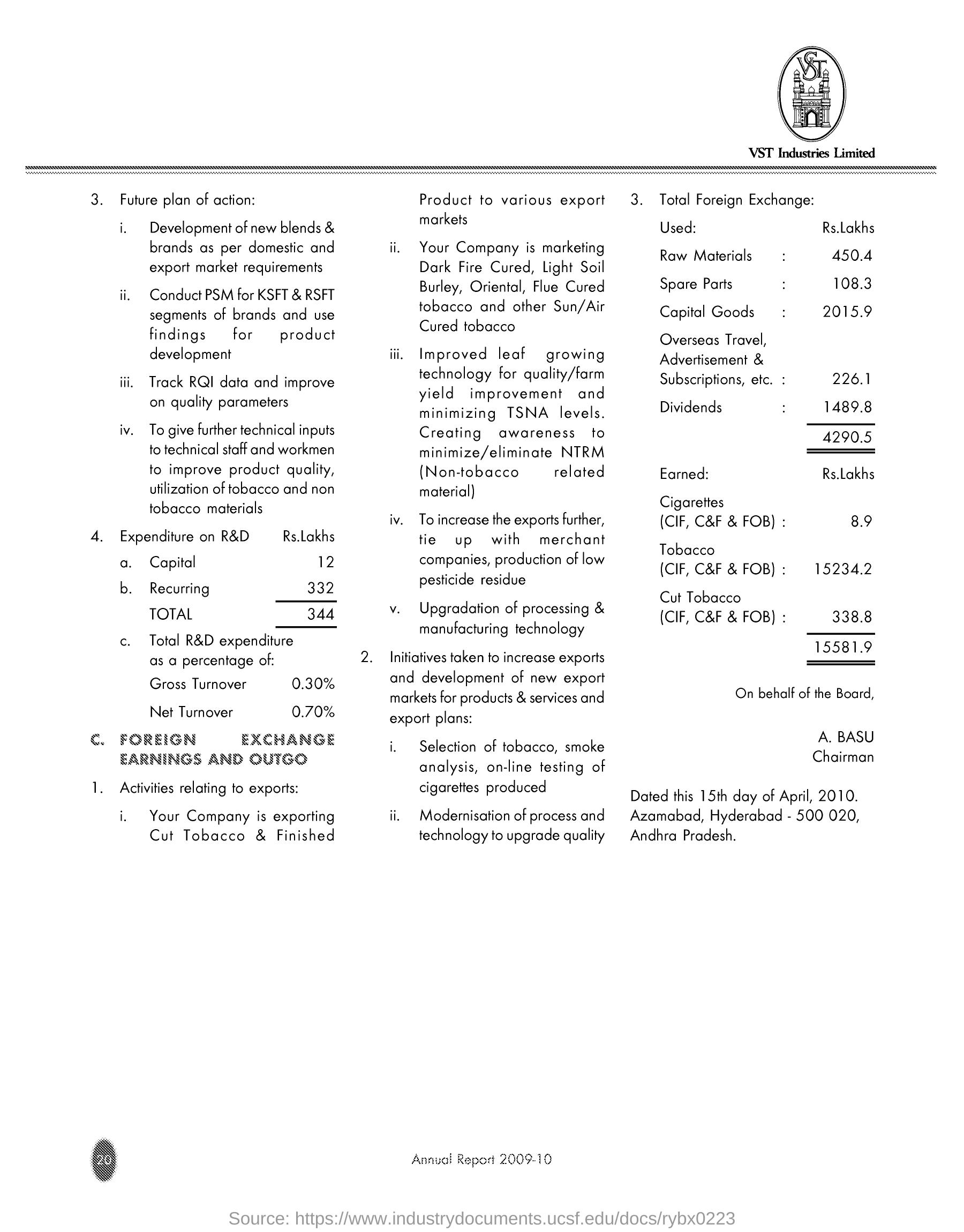} &
  \includegraphics[width=0.30\linewidth]{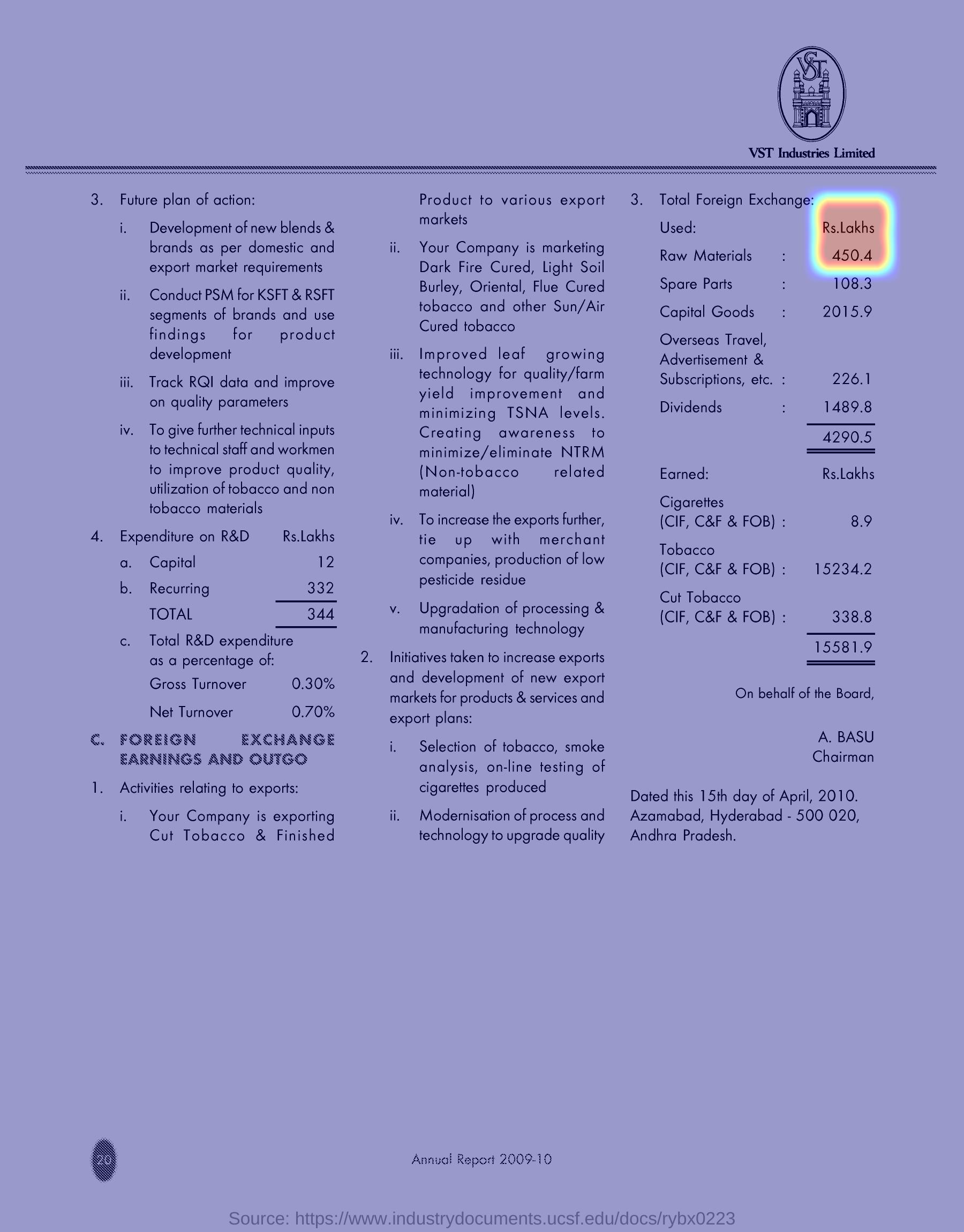}&
  \includegraphics[width=0.30\linewidth]{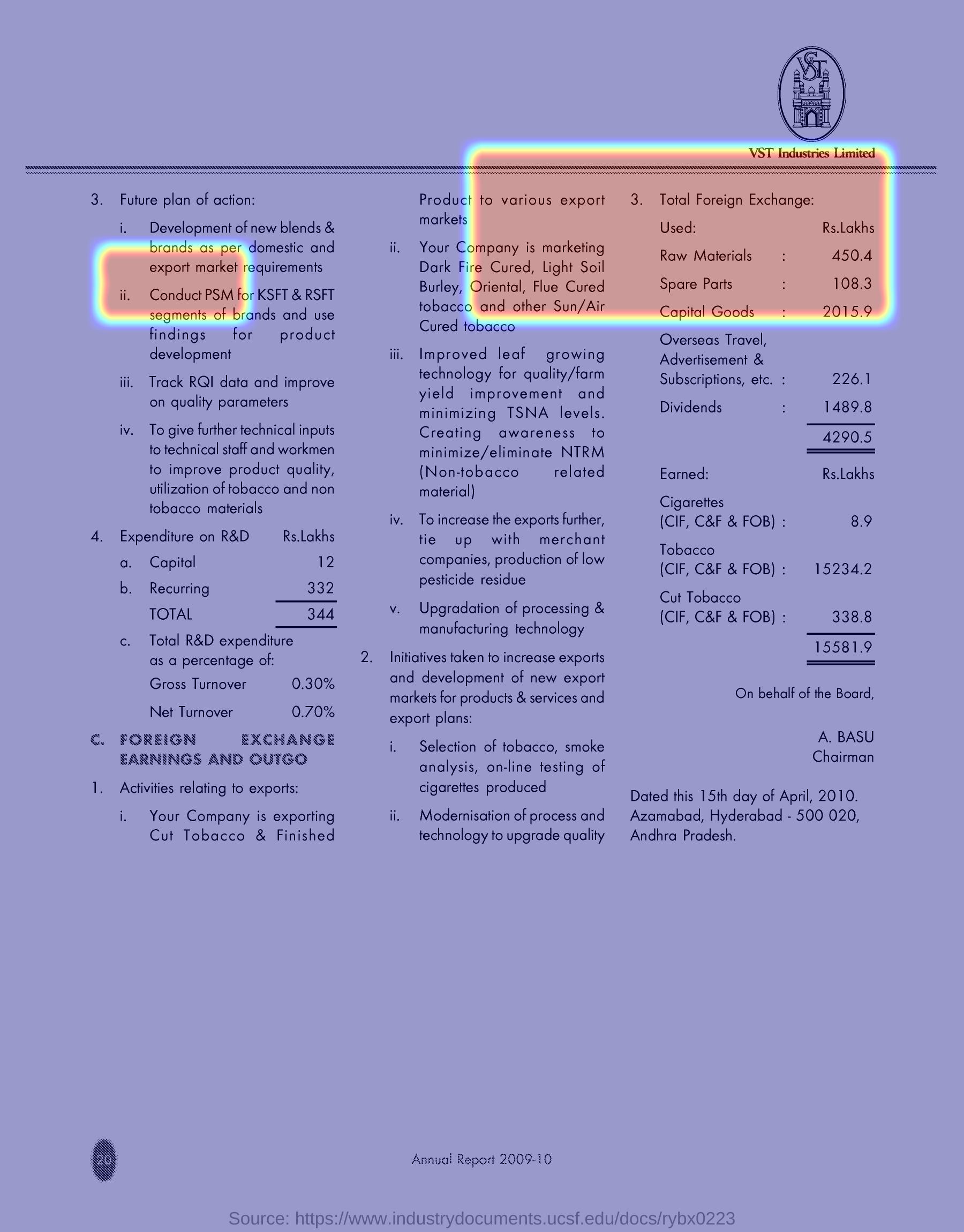} \\
  \makecell{What is the total foreign exchange\\used for raw materials(Rs.lac)??} & Clear Answer: ``450.4'' & Clear Answer:  ``450.4''
 \\ 
 Answer:  \textbf{``450.4''} & Masked Answer: ``450.4'' & Masked Answer: ``450.4''
\\ 
  \includegraphics[width=0.30\linewidth]{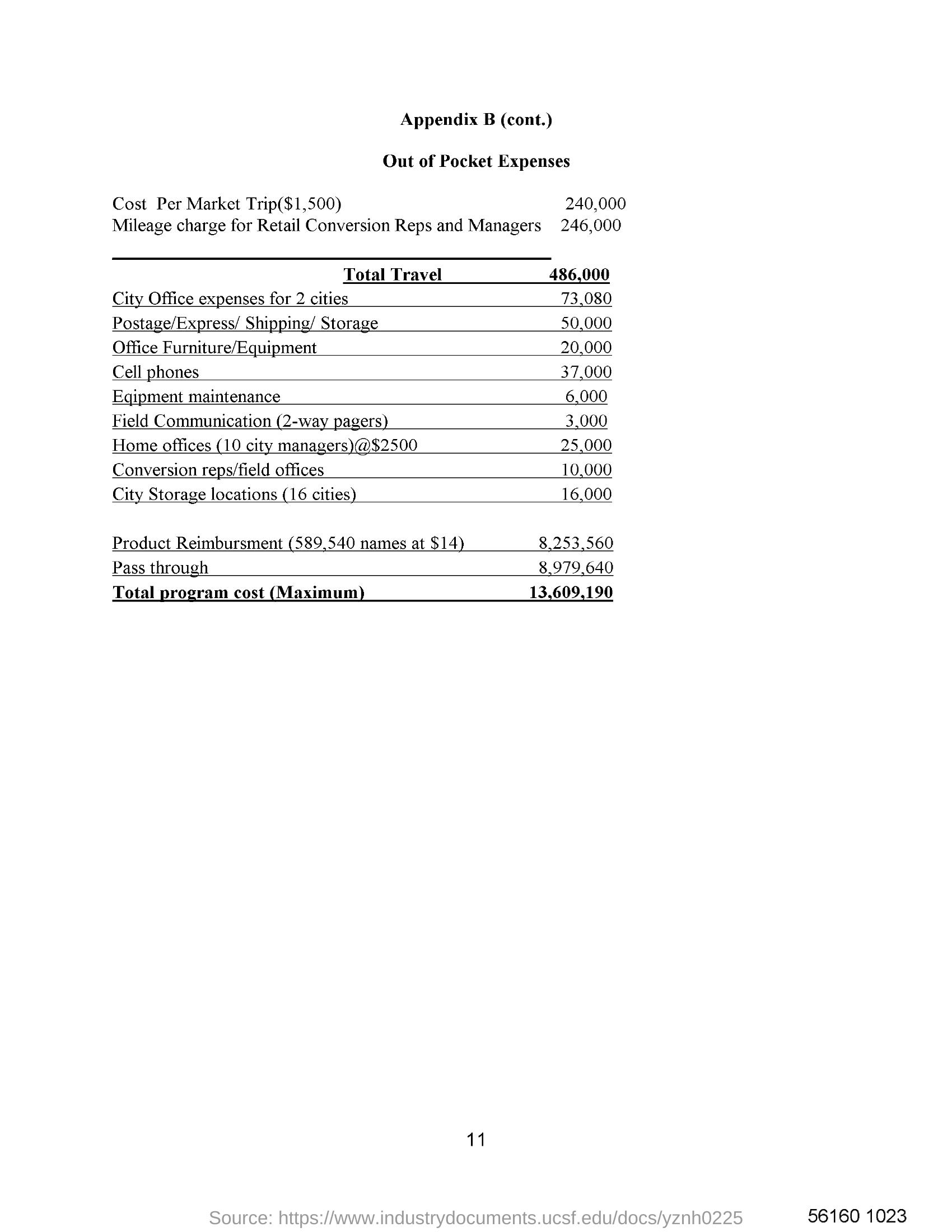} &
  \includegraphics[width=0.30\linewidth]{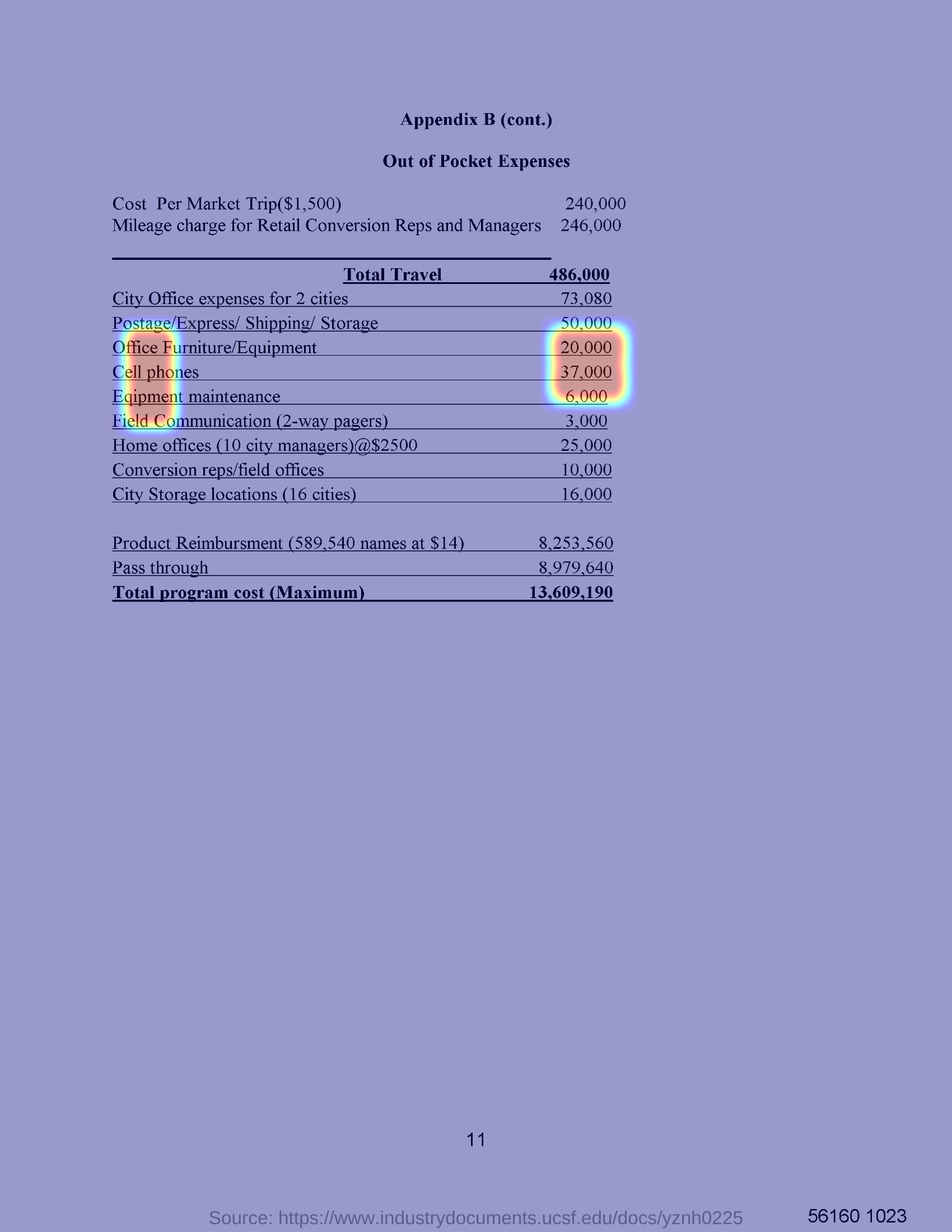}&
  \includegraphics[width=0.30\linewidth]{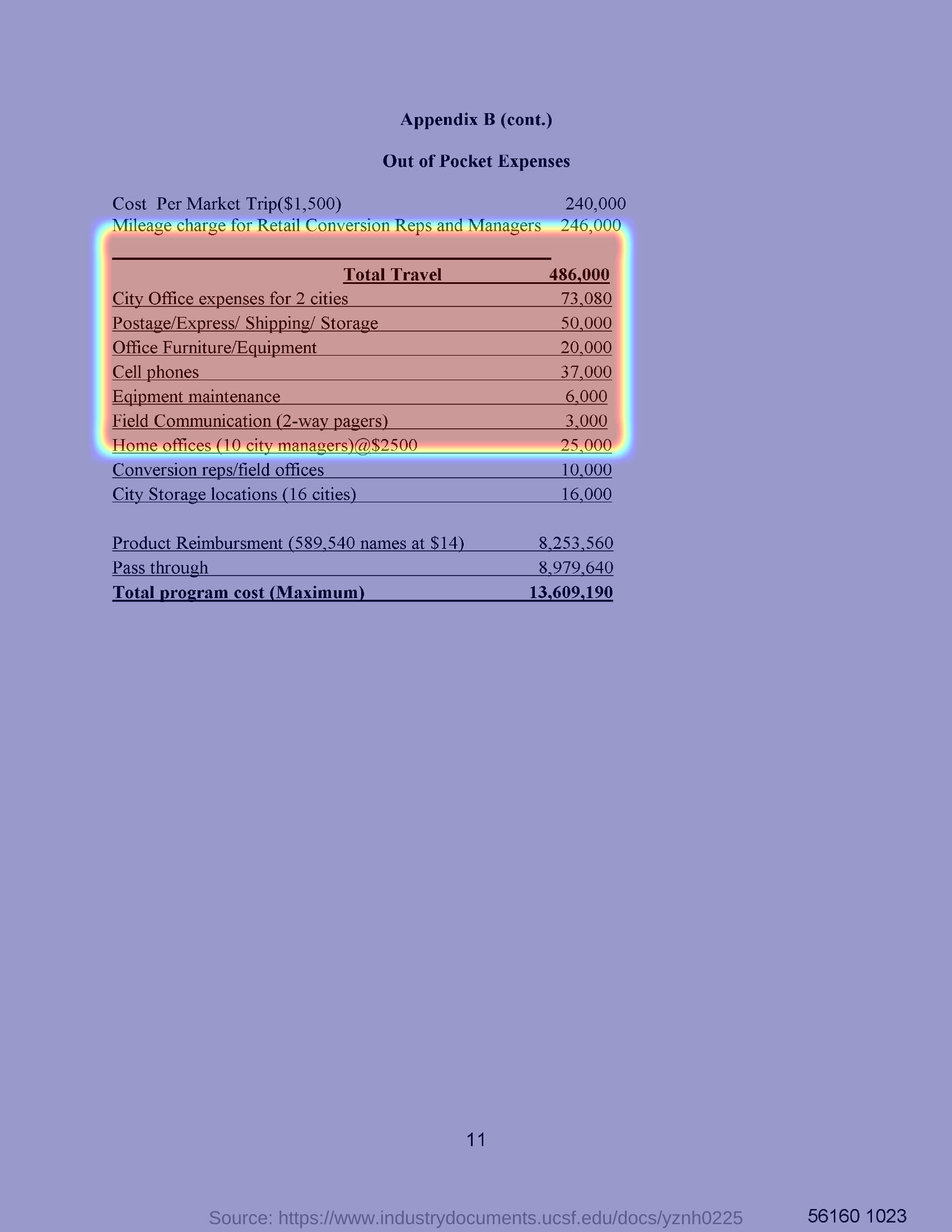} \\
  \makecell{What is the equipment\\maintenance expenses?} & Clear Answer: ``6,000'' & Clear Answer:  ``6,000''
 \\ 
 Answer:  \textbf{``6,000''} & Masked Answer: ``6,000'' & Masked Answer: ``6,000''
\\ 
\includegraphics[width=0.30\linewidth]{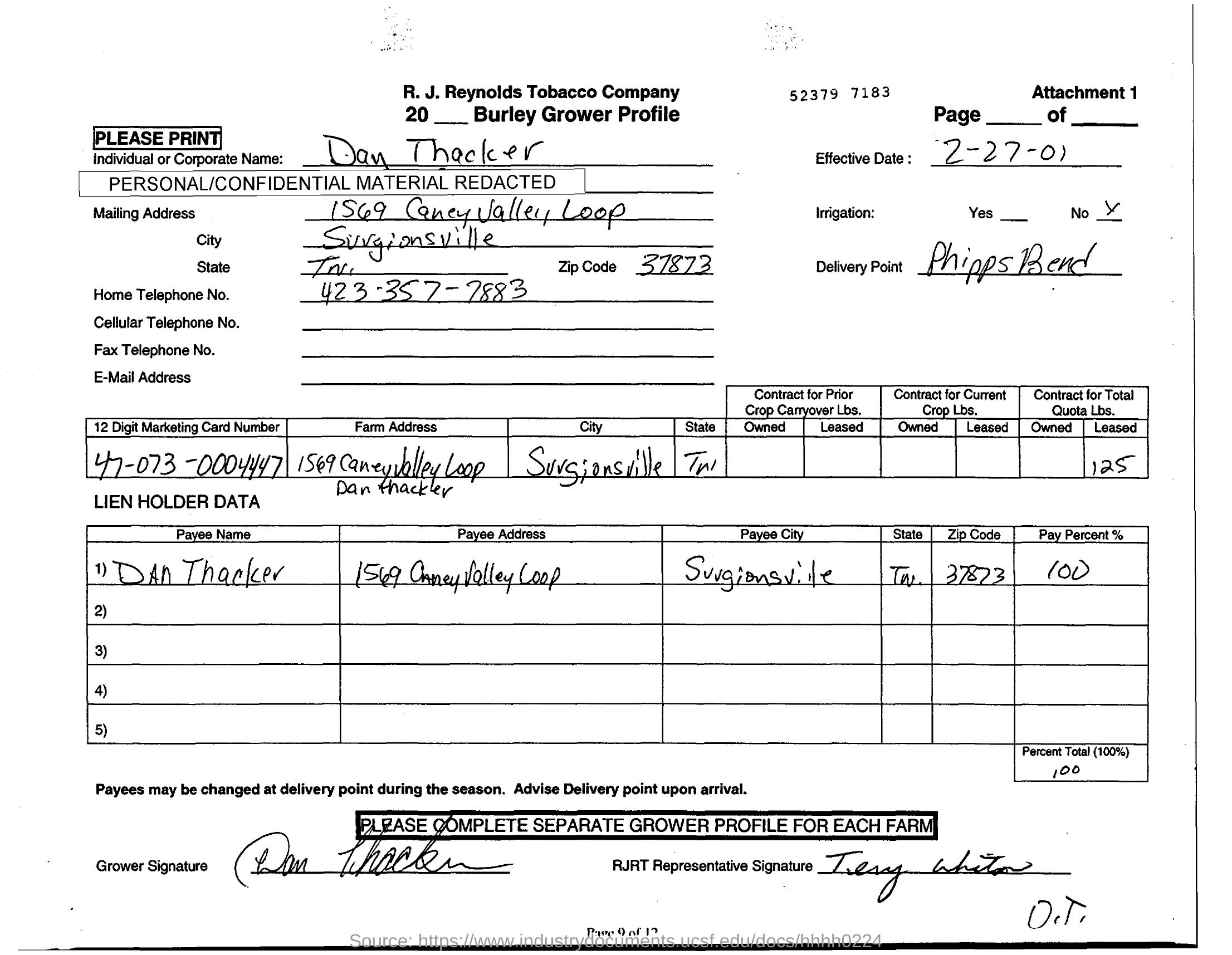} &
  \includegraphics[width=0.30\linewidth]{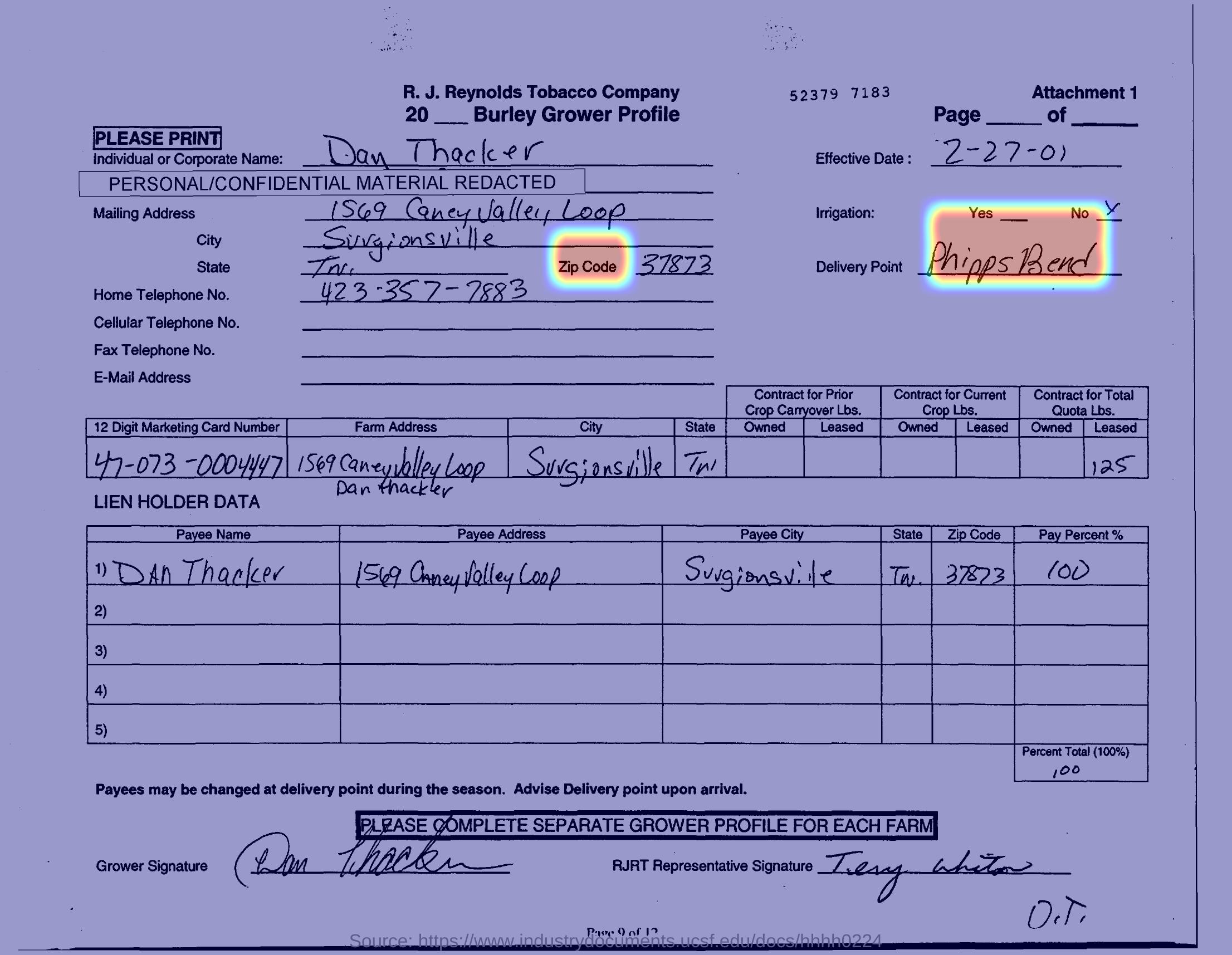}&
  \includegraphics[width=0.30\linewidth]{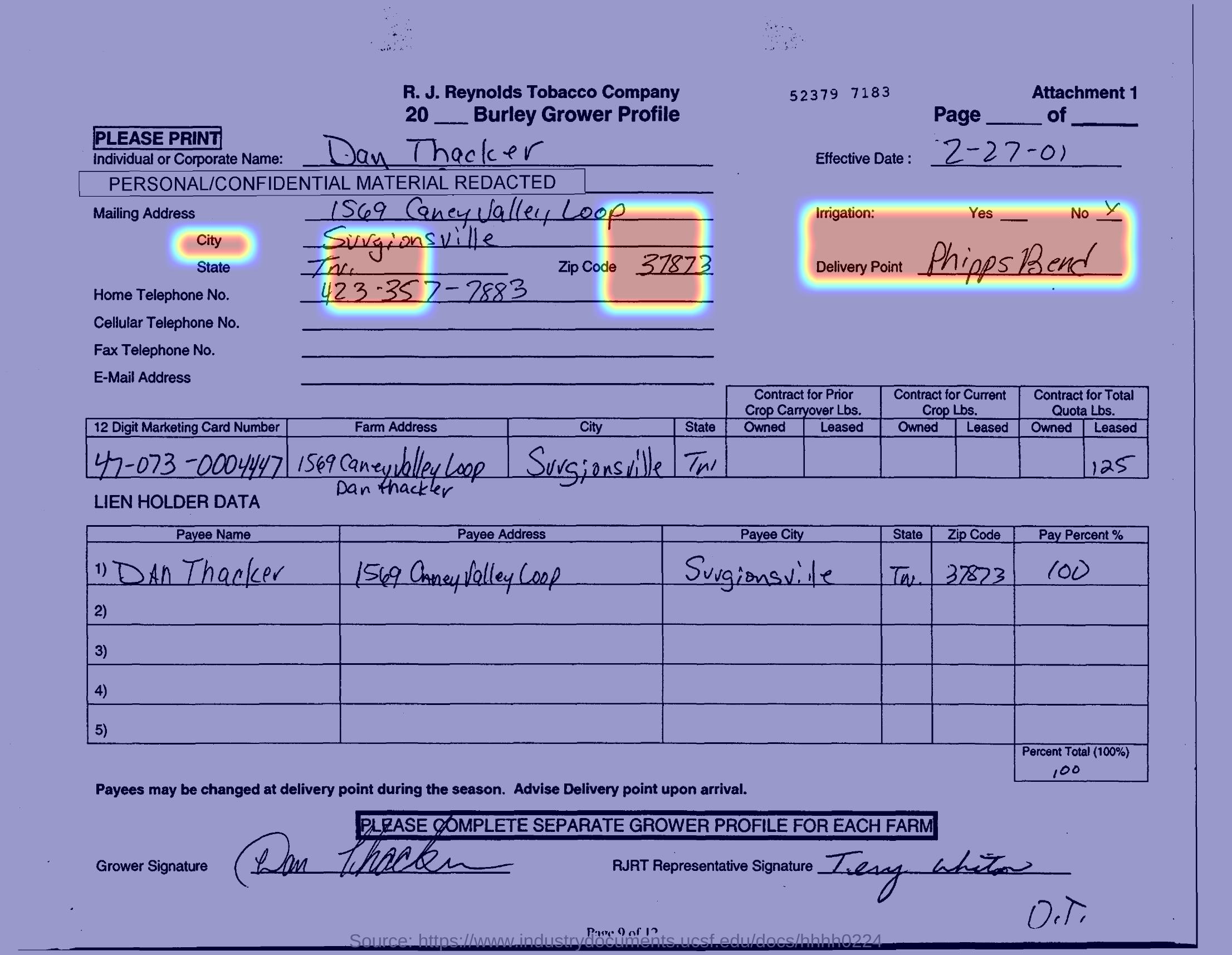} \\
  \makecell{What is the delivery point\\mentioned in the form?} & Clear Answer: ``philips bank'' & Clear Answer:  ``miops Reid''
\\ 
 GT Answer:  \textbf{``phipps bend''} & Masked Answer: ``philips bank'' & Masked Answer: ``shipping pound''
\end{tabular}
\caption{Visualization of the effect of using token interactions in our method. Left: Original Image. Middle: Masked image with relevance map learned without token interactions loss. Right: Masked image with relevance map learned with token interactions loss. }
\label{fig:interactivity1}
\end{center}
\vskip -0.2in
\end{figure}

As we discussed in \cref{sec:colpali}, training the model using the objective function in~\cref{eq:IB_basic} (without the use of token interactivity loss), lead to  overfitting to the regions that directly corresponding to the answer. This overfitting reduces the mask's utility, rendering it a mere visual reproduction of the textual answer rather than providing meaningful, context-aware insights. This can be seen in \cref{fig:interactivity1},  where we show qualitative results of our method. 

\newpage
\subsection{Additional qualitative results}
\label{sec:examples}
In this section, we present the results of the applied relevant regions masks on the images (right) across diverse contextual scenarios, in comparison to the corresponding full images (left).

\paragraph{Complex layouts} 
We test our approach on documents with complex layouts from the InfographicVQA dataset~\cite{mathew2022infographicvqa}. 

\begin{figure}[h] 
    \centering
    \includegraphics[width=0.7\linewidth]{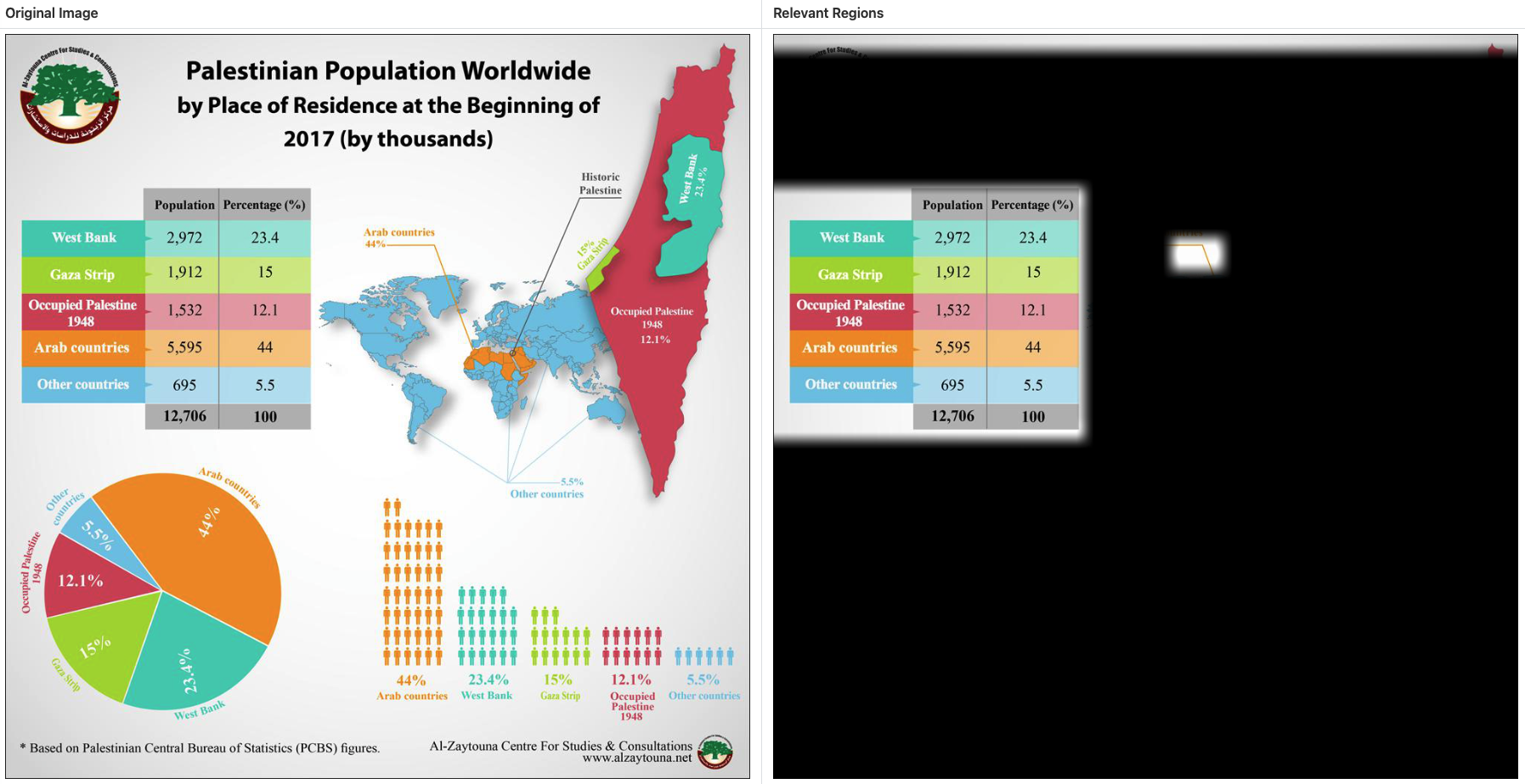} 
    \caption{\textbf{Question:} What is the total percentage of Palestinians residing at places other than West Bank and Arab countries? \\\textbf{Answer:} 32.6 \% (correct).} 
    \label{fig:palestinian}
\end{figure}

\begin{figure}[h]
    \centering
    \includegraphics[width=0.7\linewidth]{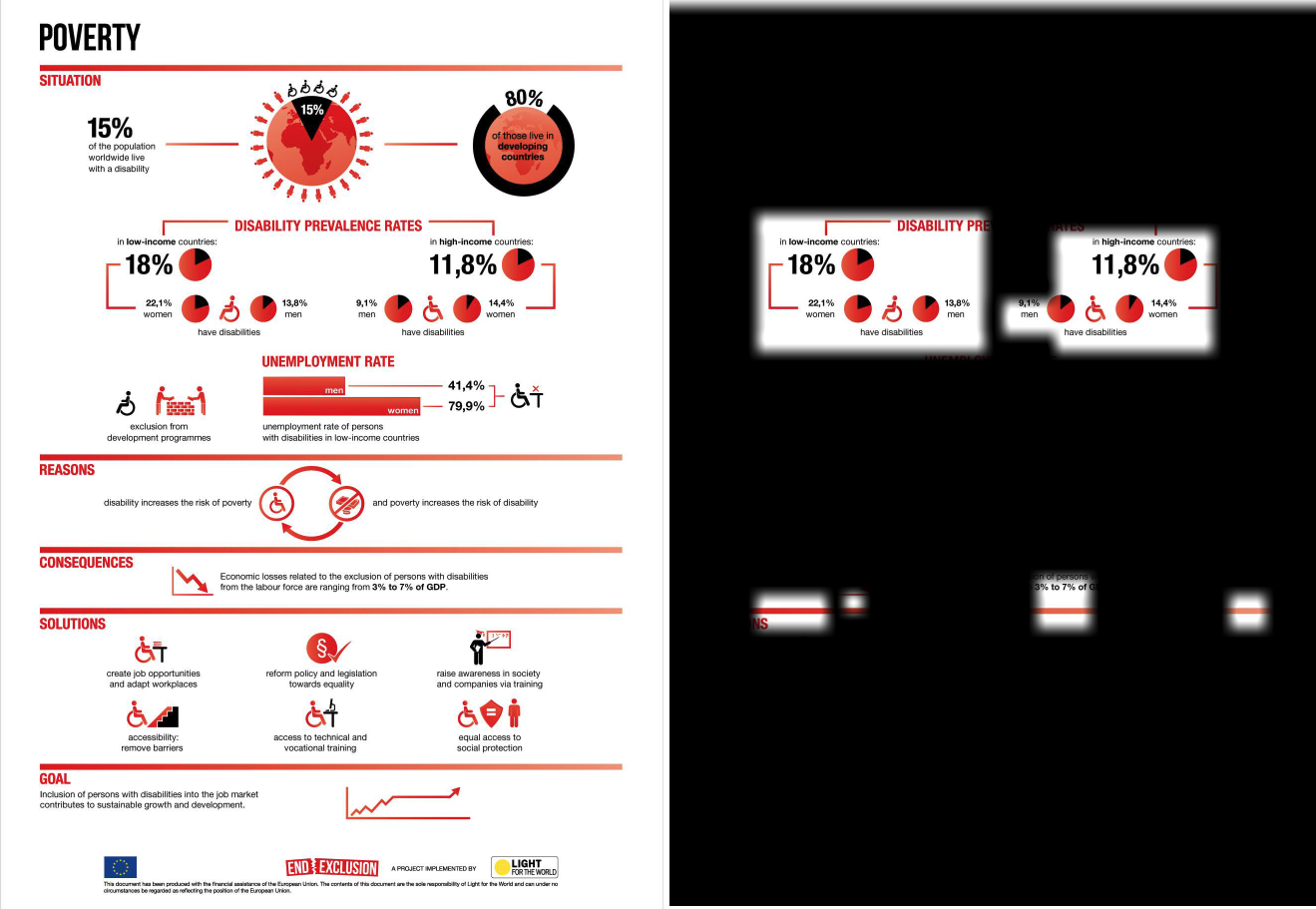}
    \caption{\textbf{Question:} What is the percentage of women with disabilities in low income countries? \\\textbf{Answer:} 22.1 \% (correct).} 
    \label{fig:poverty}
\end{figure}

\newpage
\paragraph{Densely populated documents} We present results on documents containing dense textual content.

\begin{figure}[h] 
    \centering
    \includegraphics[width=0.70\linewidth]{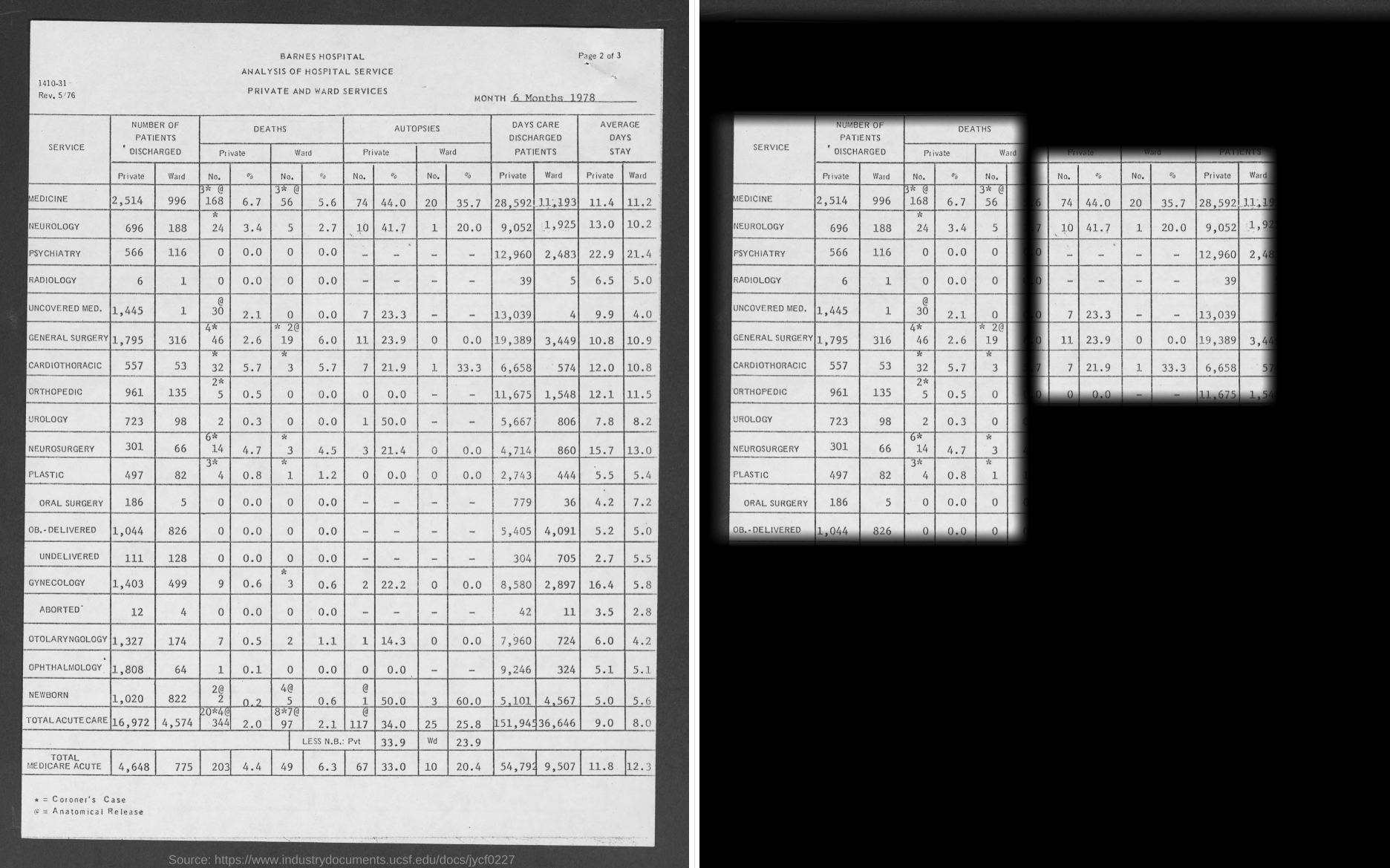} 
    \caption{\textbf{Question:} Under Private service how many patients were discharged in Neurology? \\\textbf{Answer:} 696 (correct).} 
    \label{fig:table}
\end{figure}

\begin{figure}[h] 
    \centering
    \includegraphics[width=0.70\linewidth]{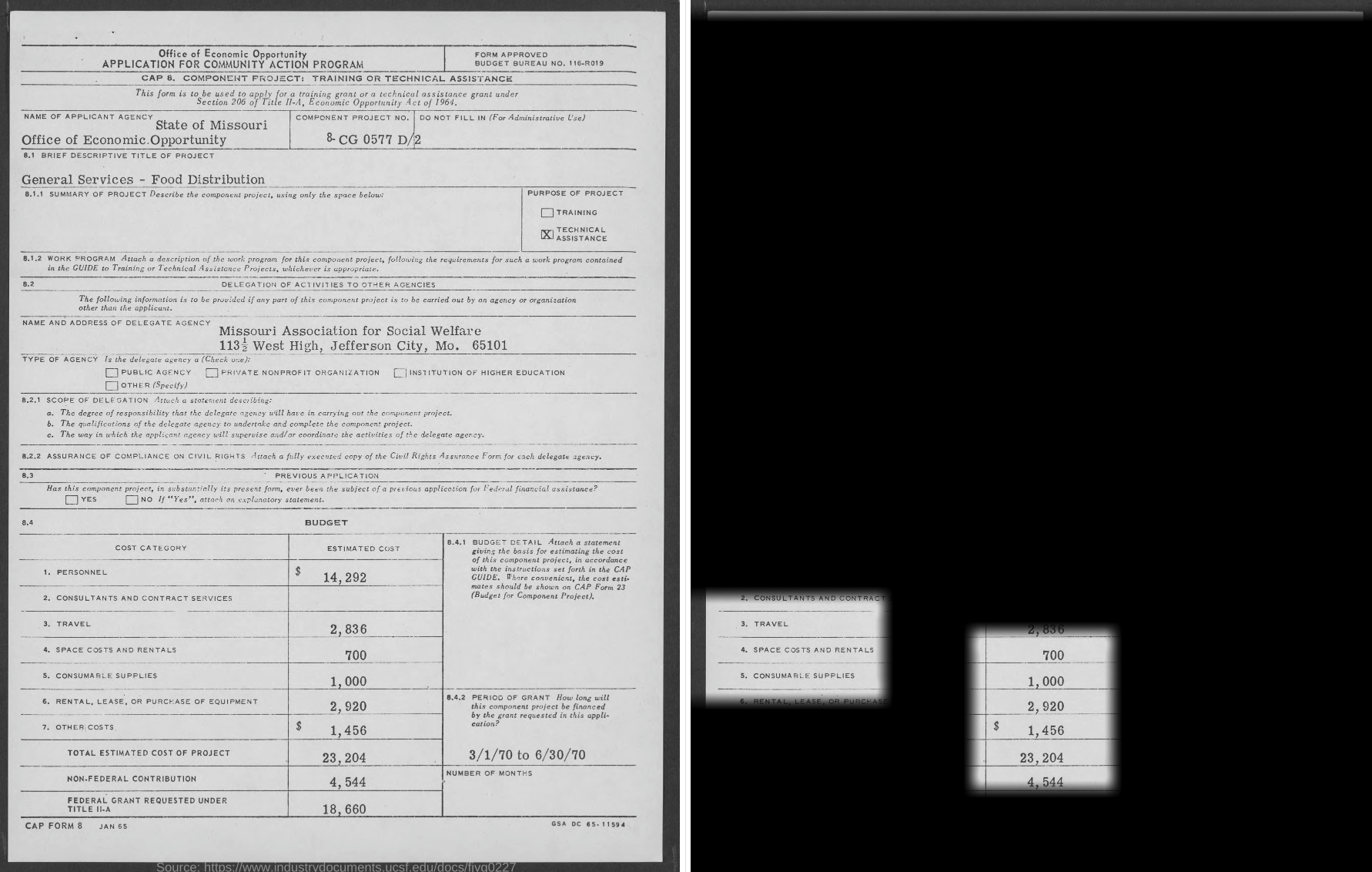} 
    \caption{\textbf{Question:} What is the estimated cost of Space costs and Rentals? \\\textbf{Answer:} 700 (correct).} 
    \label{fig:economy}
\end{figure}
\newpage
\paragraph{Degraded quality} We present results on documents with degraded visual quality.

\begin{figure}[h] 
    \centering
    \includegraphics[width=0.7\linewidth]{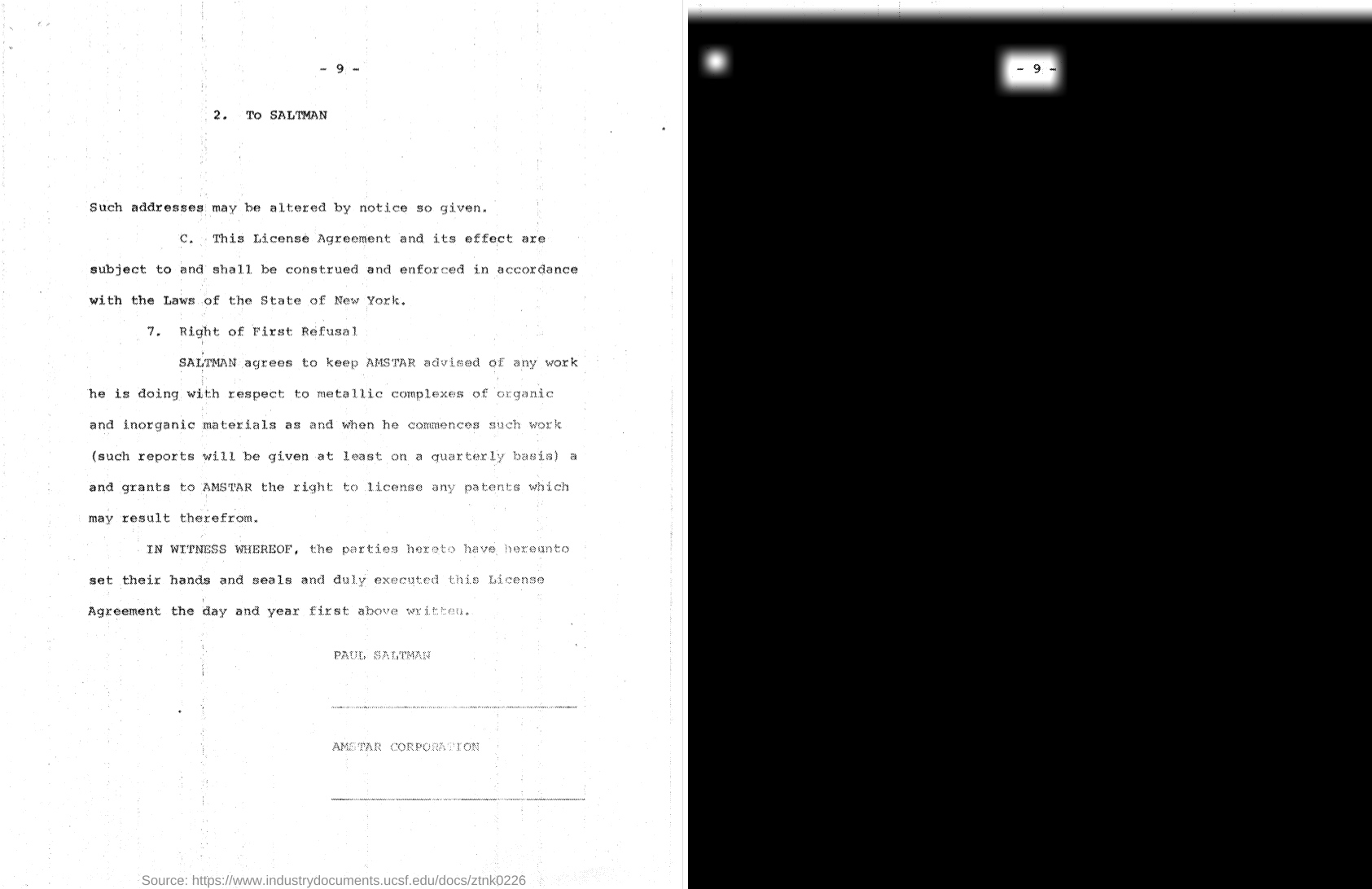} 
    \caption{\textbf{Question:} What is the page number? \\\textbf{Answer:} 9 (correct).} 
    \label{fig:degraded}
\end{figure}

\begin{figure}[h] 
    \centering
    \includegraphics[width=0.7\linewidth]{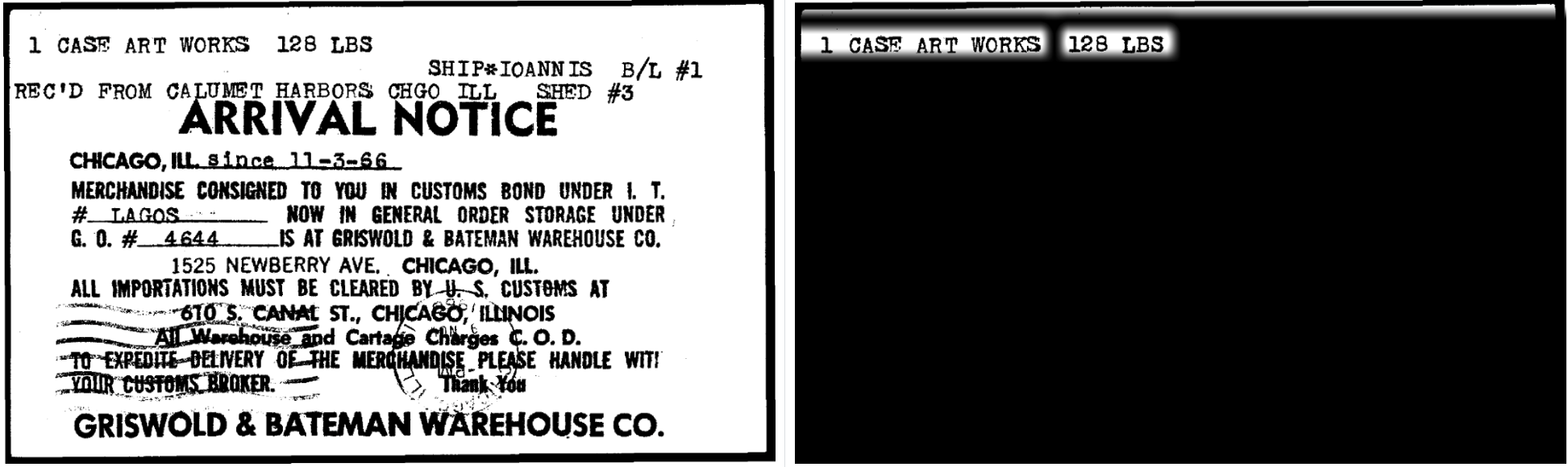} 
    \caption{\textbf{Question:} How many cases of artworks are there in the shipment? \\\textbf{Answer:} 1 (correct).} 
    \label{fig:shipment}
\end{figure}

\begin{figure}[h] 
    \centering
    \includegraphics[width=0.7\linewidth]{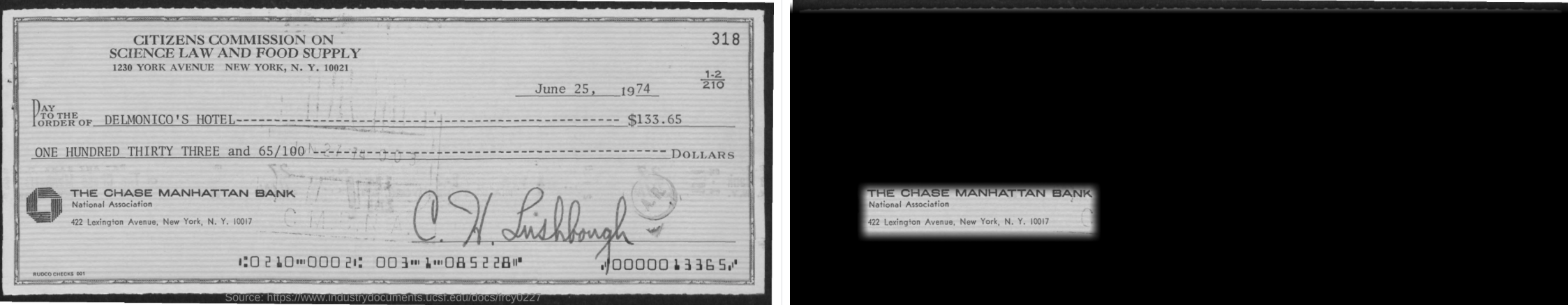} 
    \caption{\textbf{Question:} Which bank's check is this? \\\textbf{Answer:} The chase manhattan bank (correct).} 
    \label{fig:foodsupply}
\end{figure}
\newpage

\subsection{Results with a different backbone}
\label{app:sec:donut}
\begin{figure}[h] 
    \centering
    \includegraphics[width=0.95\linewidth]{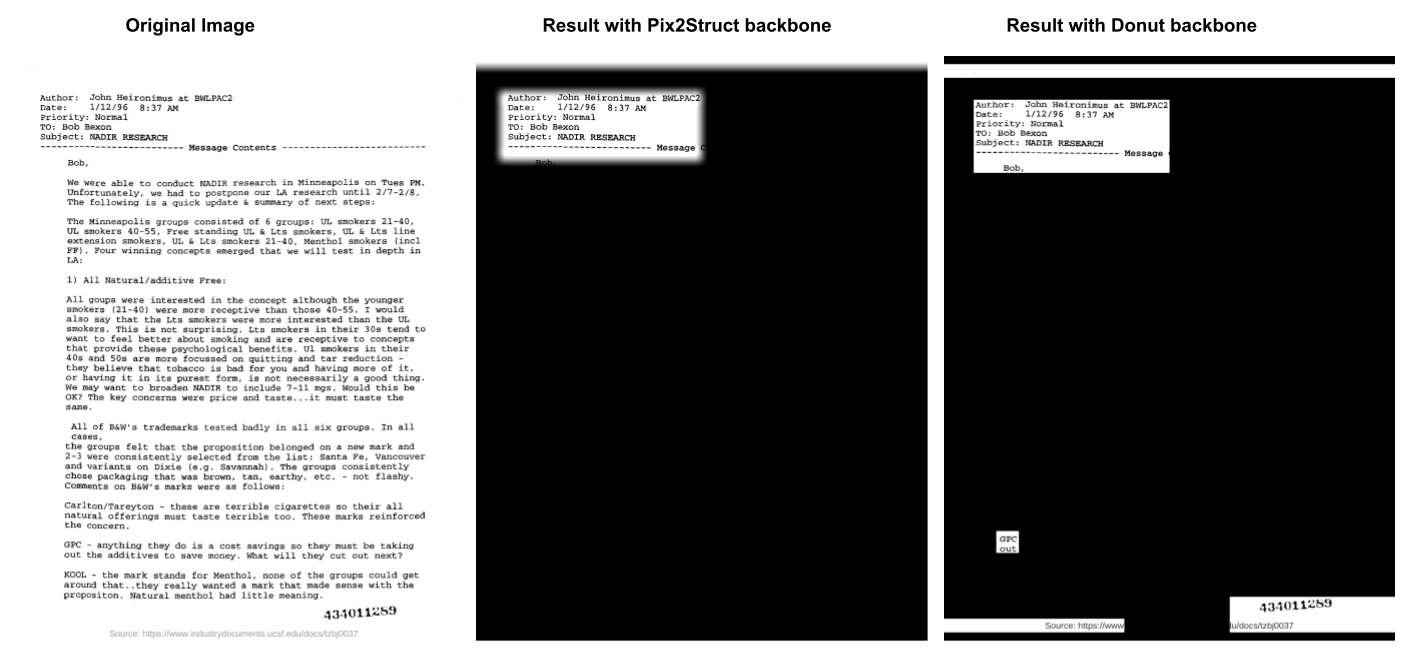} 
    \caption{\textbf{Question:} Who is this question addressed to?\\\textbf{Answer:} bob bexon (correct).} 
    \label{fig:b4-address}
\end{figure}

\begin{figure}[h] 
    \centering
    \includegraphics[width=0.95\linewidth]{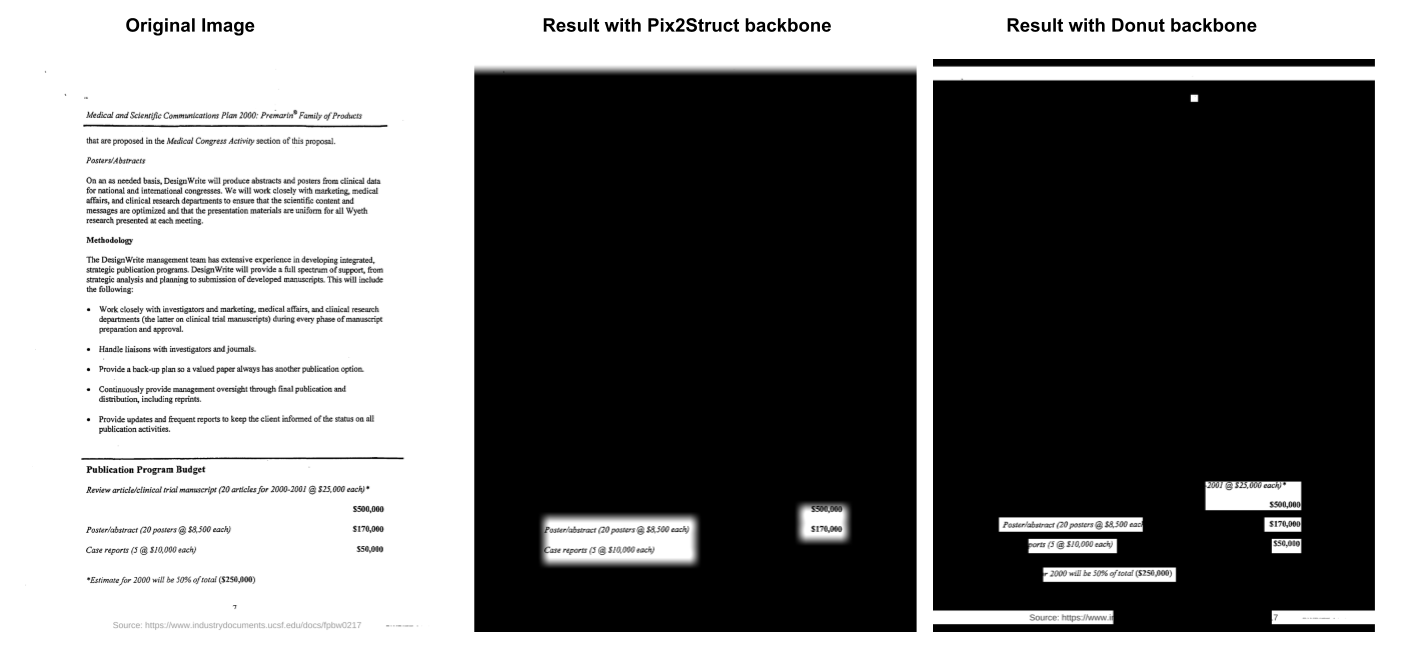} 
    \caption{\textbf{Question:} What is the amount for publising one poster / abstract?\\\textbf{Answer:} \$8,500 (correct).} 
    \label{fig:b4-posters}
\end{figure}

\begin{figure}[h] 
    \centering
    \includegraphics[width=0.95\linewidth]{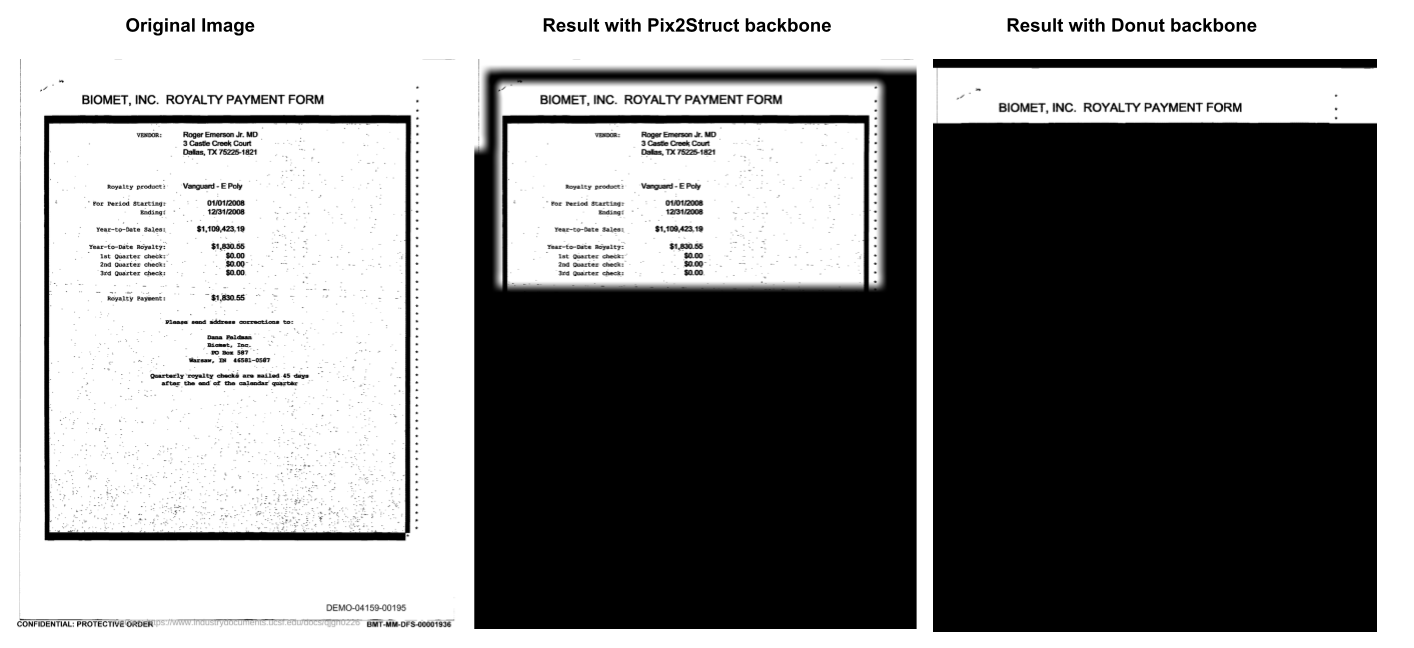} 
    \caption{\textbf{Question:} what type of communication is issued by Biomet, inc.? \\\textbf{Answer:} royalty payment form (correct).} 
    \label{fig:b4-payment}
\end{figure}

In this section, we show the results of \MET using the Donut backbone~\cite{kim2021donut}, and compare it with the Pix2Struct backbone. The obtained results are given in \cref{fig:b4-address}, \cref{fig:b4-posters} and \cref{fig:b4-payment}, demonstrating the model-agnostic capability of our method. These figures highlight the relevant regions produced by \MET using two different backbones while answering the same questions.

\end{document}